\documentclass{aimsessay}
\usepackage[utf8]{inputenc}
\usepackage[round]{natbib}
\usepackage{amsmath}
\usepackage{amsfonts}
\usepackage{amssymb}
\usepackage{mathtools}
\usepackage{latexsym}
\usepackage{parskip}
\usepackage{enumerate}
\usepackage{tikz}
\usepackage{caption} 
\usepackage{subcaption} 
\usepackage{arabtex}
\usepackage{utf8}
\setcode{utf8}
\usepackage{booktabs} 
\usepackage{array}    
\usepackage[below,section]{placeins} 
\usepackage{moreverb}   
\swapnumbers 
\theoremstyle{plain}
\newtheorem{thm}[subsection]{Theorem}
\theoremstyle{definition} 
\newtheorem{lem}[subsection]{Lemma}

\newtheorem{exa}[subsection]{Example}

\newtheorem{rem}[subsection]{Remark}
\numberwithin{equation}{section}
\title{Machine Learning Methods for Shark Detection}
\author{Jordan Felicien Masakuna (jordan@aims.ac.za)\\
African Institute for Mathematical Sciences (AIMS)\\
\\
{\small Supervised by: Dr.\,Simukai Utete and Prof.\,Jeff Sanders} \\
{\small AIMS, South Africa}%
}
\date{{\small 21 May 2015}\\%
  {\scriptsize\it Submitted in partial fulfillment of 
    a structured masters degree at AIMS South Africa}\\%
  \vspace{0.5cm}{\includegraphics{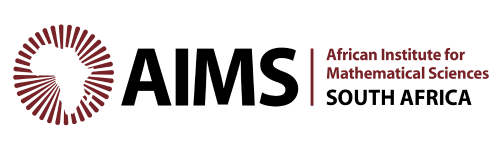}}}
\begin{document}
\pagestyle{empty}
\maketitle
%
\pagenumbering{roman}
\chapter*{Abstract} 
\addcontentsline{toc}{chapter}{Abstract}
This essay reviews human observer-based methods employed in shark spotting in Muizenberg Beach. It investigates Machine Learning methods for automated shark detection with the aim of enhancing human observation. A questionnaire and interview were used to collect information about shark spotting, the motivation of the actual Shark Spotter program and its limitations. We have defined a list of desirable properties for our model and chosen the adequate mathematical techniques. The preliminary results of the research show that we can expect to extract useful information from shark images despite the geometric transformations that sharks perform, its features do not change. To conclude, we have partially implemented our model; 
the remaining implementation requires dataset. 

\textbf{Key words:} Feature Extraction, Classification, Fusion, Support Vector Machine, $k-$Nearest Neighbours, Neural Network, Dempster-Shafer Fusion, Fourier Descriptor, Legendre Moment, Complex Moment.


\vfill
\section*{Declaration}
I, the undersigned, hereby declare that the work contained in this research project is my original work, and that any work done by others or by myself previously has been acknowledged and referenced accordingly.

\includegraphics[height=2cm]{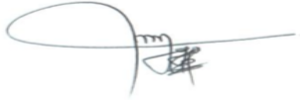} \newline \hrule
Jordan Felicien Masakuna, 21 May 2015

\tableofcontents
\newpage
%
\pagenumbering{arabic}
\pagestyle{myheadings}
\chapter{Introduction}

\begin{figure}[h]
   \centering
    \begin{subfigure}[b]{0.34\textwidth}
        \centering
        \includegraphics[width=\textwidth]{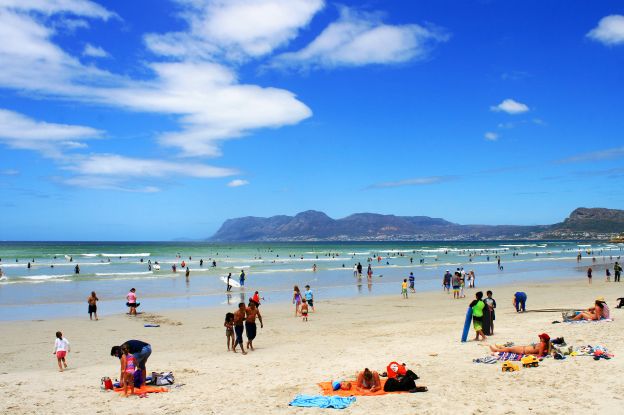}
        \caption{Muizenberg Beach \href{http://www.surfers-corner.co.za/2012/12/30/sunny-muizenberg-beach/}{http://www.surfers-corner.co.za}}
        \label{fig:probl1}
    \end{subfigure}
   \begin{subfigure}[b]{0.34\textwidth}
        \centering
        \includegraphics[width=\textwidth]{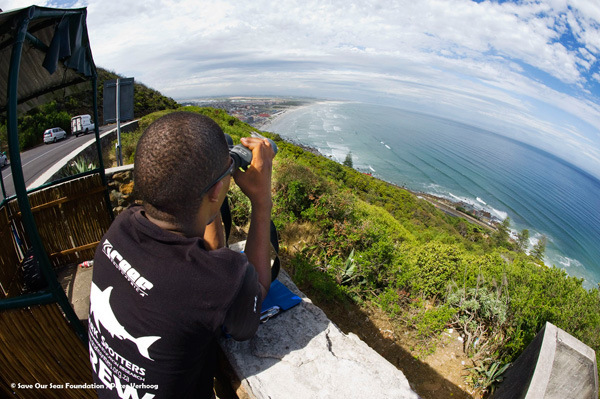}
        \caption{shark spotter \href{http://sharkspotters.org.za/about/photos}{http://sharkspotters.org.za}}
        \label{fig:probl2}
    \end{subfigure}
    \caption{shark spotter}
    \label{fig:problem}
\end{figure}

Many Cape Town people spend their holidays on beaches either swimming or surfing. There is danger in the water with the presence of sharks which can cause death. 
Nevertheless, people come and enjoy themselves at the beaches as shown in Figure \ref{fig:probl1}. Great white shark is the type of shark most spotted in Cape Town and it can reach 6.1 meters of length. Naturally in Summer, sharks can frequent beaches (\href{http://sharkspotters.org.za/information}{http://sharkspotters.org.za/information}).

\begin{figure}[h]
    \centering
        \includegraphics[scale=0.5]{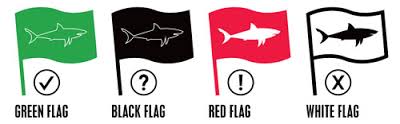}
    \caption{Flags (\href{http://sharkspotters.org.za/how-it-works/flag-system-protocol}{http://sharkspotters.org.za/})}
    \label{fig:flag}
\end{figure}

In 2004 a program named Sharks Spotters (\href{http://sharkspotters.org.za/}{http://sharkspotters.org.za/}) was formed with the objective of spotting sharks in the ocean while people are surfing. In Muizenberg for example, the spotters (see Figure \ref{fig:probl2}) are positioned on the mountain and scan a specific area of the water through binoculars. The spotters on the mountain are in communication with other spotters on the beach. There are four types of flags (see Figure \ref{fig:flag}) that spotters use to communicate with other spotters, surfers and swimmers. The four flags transmit the following messages: (1) \textit{Green flag}: the conditions are good for spotters to work. (2) \textit{Black flag}: the conditions are poor for spotters to perform well. (3) \textit{Red flag}: be warned a shark is in the water. (4) Finally, the \textit{White flag} signals surfers and swimmers to leave the water immediately in order to avoid the risk of death.


When we interviewed the spotters they mentioned that poor conditions such as rain, murky or rough weather, cloudy and windy conditions make it difficult to spot sharks. 
Consequently, spotters run the two following risks: (1) \textbf{False Positive}: spotters could ask surfers to evacuate under the false belief that a shark has been spotted. 
If this occurs frequently, surfers will no longer trust spotters. This is the first danger. (2) \textbf{False Negative}: spotters might not see a shark while there is one in the water. This is the second danger. The purpose of this work is thus to propose an automatised system supporting decision-making when a suspect object is spotted. An image is taken of a suspect object and both humans and an automated system classify the object whether it is a shark or not, compare the results and make a decision. We need to reduce the risks of False Positive and False Negative by converting the actual manual shark spotter program to a semi-automated shark spotter.

Shark spotting involves three main steps: $(1)$ detecting a suspect shape in the water; $(2)$ deciding whether or not it is a shark; $(3)$ delivering a message to surfers when there is a shark. The automated shark spotter that we propose involves the three steps above stated as well. We aim to design shark detection that is able to: (1) detect sharks despite the geometric transformations they can perform; (2) detect sharks despite poor weather conditions. 

In order to fulfil those aims, we consider shark spotting as a multi-class classification task in Machine Learning. We need to classify four objects, namely a mature shark, a school of sharks, a baby shark and other objects in the water. A classification assigns a new image to one of the four classes. After spotting a school of sharks, the spotter has to communicate with the nearest beach and the surrounding beaches as well. 
The risk of baby sharks is less considerable than the risk of mature sharks. Since the actual program gives good results apart from when poor weather conditions prevail, we need to design a model that can support the existing system. \cite{DSF1} considered how to fuse different classification techniques in order to improve their outputs. We also consider different classification techniques and fuse their outputs. 

Before classifying images, we prepare them by reducing noise and improving image contrast caused by poor weather, detect different shapes and extract useful information. We assume that data is not linearly separable, meaning it is not possible to separate the four classes linearly in space.  And the image to be classified contains objects in the ocean such as fish, people and water.

To extract useful information from an image we focus on obtaining space-based and frequency-based features. We combine space-based and frequency-based outputs to diversify the features. We use a Generic Fourier Descriptor during frequency-based feature extraction. We use a Complex Moment and a Legendre Moment for space-based feature extraction. The choice of the two space-based features is motivated by the fact that one constructs an orthogonal set of features while the other does not \citep{GM}.

To classify an image, we first use an Artificial Neural Network (ANN) which trains a classifier from training data so that it can assign a new image to the class using its knowledge. ANN evaluates the training error based on inputs and minimizes the error \citep{ANN1}. 
Then $k$-Nearest Neighbours which evaluates the closeness between the query image and every image in data set, looks for $k$ images that are very close to the query image and proceeds by voting. A query image $x_q$ is assigned to class $c_k$ having more images among $k$ images chosen \citep{KNN1}.
Then a Support Vector Machine (SVM) reduces the classification error. SVM transforms a non-linear problem to a linear problem by using the kernel function that converges to a global minimal since the problem is converted to a convex problem \citep{SVM1}. We use Dempster-Shafer Fusion (DSF) to fuse classification technique outputs. DSF considers the uncertainties of data and combines different sources of evidence \citep{DSF2}.

The rest of the essay has the following organization: 
Chapter 2 will present a list of model properties in order to get a good output and decide how to choose the mathematical techniques to be used. Chapter 3 will show the different preliminary tasks applied to an image in order to reduce noise, to improve its contrast, to segment an image and to extract useful information. Chapter 4 will present the three classification techniques to be used and how they are combined in order to improve their outputs. Chapter 5 presents the results obtained and a discussion thereof. In Chapter 6, we will conclude and present further work. 
\chapter{Model Properties}
We are concerned with the classification task in Machine Learning that assigns a query image to one of the four classes. The problem that can arise is that two classifiers can classify the same query image differently even though they are using the same training data. We mitigate this classification error by defining the desired properties that make a good classifier. The purpose of a classifier is to find a function $F(x_i)=F(\mathcal{X},x_i)$ which behaves like the exact unknown function $\mathcal{F}(x_i)$. The function $\mathcal{F}(x_i)$ is defined as follows:
\begin{equation}
\mathcal{F} :  \mathcal{X} \to \Omega\,, \label{eq:class1}
\end{equation}
where $\mathcal{X}$ and $x_i$ are the set of training data and its $i^\text{th}$ image and $\Omega=\{c_1,\cdots,c_k\}$ is the set of classes. In the next section we define a list of properties that the system should have in order to ensure that accurate results are obtained from the shark detection model, i.e.\,to prove the accuracy of the model.
\section{Properties}
A good  model has to cover the following properties that represent the relationship between the input $\mathcal{X}$ and the output $\mathcal{F}(x_i)$.  This list of properties is not exhaustive, but it is enough to design a model that satisfies the requirements of shark detection. The notation to be used throughout this work is as follows. Let 
\begin{itemize}
\item $\mathcal{X} \in \mathbb{R}^{n \times m}$ and $Y\in \Omega^n$ be the sequences of training data and outputs respectively;
\item $\mathcal{X}^\circ$ and $\mathcal{X}^c \subset \mathcal{X}$ be the permutation of two rows ($i^{th}$ and $j^{th}$) of $\mathcal{X}$ and the features set respectively;
\item $x_i \in \mathcal{X}$, $y_i \in Y$ and $x_q$ be the $i^\text{th}$ image in $\mathcal{X}$, its output and the query image respectively;
\item $R(\mathcal{X})$ and $N(\mathcal{X})$ be the number of images in $\mathcal{X}$ and the number of pixels in $x_i$ respectively;
\item $\mathcal{M}=\{M_1,\cdots, M_l\}$ be the set of classifier models;
\item $\Gamma(f,h)$ be the error between two functions $f$ and $h$;
\item $g(x,y)$ be the grayscale intensity where $(x,y)$ represents its location;
\item $w(x_q) \in [0,1]^{l \times k}$ be the decision profile matrix such that $\sum_{j=1}^k w_i^j(x_q) =1$.
\end{itemize}
\textbf{Property 1. (Optimality)} The optimal function of $F(x_i)$ is the exact unknown function $\mathcal{F}(x_i)$. It is easy to get the exact function, i.e.\,$F(x_i)=\mathcal{F}(x_i)$ when we deal with linear separable data, but it is difficult for non-linear-separable data in which we seek to minimize error between the two functions.

\textbf{Property 2. (Computational Cost)} Let $\mathcal{X}$, $\mathcal{X}^c$, $N$ and $\Gamma$ be as defined above.
\begin{equation}
N(\mathcal{X})\geq N(\mathcal{X}^c) \implies \Gamma[\mathcal{F}(\mathcal{X}^c,x_q)-F(\mathcal{X}^c,x_q)] \leq \Gamma[\mathcal{F}(\mathcal{X},x_q)-F(\mathcal{X},x_q)]\,. \label{eq:p2}
\end{equation}
\textbf{Property 3. (Graph)} The set $\mathcal{X}$ is a disconnected graph containing $k$ connected components. Images are assigned to each component based on their similarity (closeness).

\textbf{Property 4. (Misclassification)} Let $\mathcal{F}(x_i)=c_j$. A misclassification is a case where an image of class $c_j$ is assigned to class $c_k$ with $j\neq k$. $F$ is a misclassifier if and only if $ F(x_i) \neq \mathcal{F} (x_i)$ for some $x_i$.

\textbf{Property 5. (Convergence)} The classifier $F(x_i)$ fits $\mathcal{F} (x_i)$, i.e.\,it behaves like $\mathcal{F} (x_i)$ for all images in domain $\mathcal{X}$ . The error has to be minimized such that:
\begin{equation}
\Gamma[\mathcal{F} (x_i)-F(x_i)]\approx 0\,. \label{eq:p5}
\end{equation}
\textbf{Property 6. (Uniqueness)} The function classifier $F (x_i)$ is not unique and does not have an explicit algebraic expression because its form depends on the technique used. The reliability of a technique in comparison with another is not absolute, i.e.\,it depends on the data. Let $\mathcal{X}$ be as defined above, $F_1 (x_i)$, $F_2 (x_i)$ be two classifiers. Then
\begin{equation}
F_1 \;\text{is better than}\; F_2 \;\text{in}\; \mathcal{X} \iff \Gamma[\mathcal{F}(x)-F_1(x)] \leq \Gamma[\mathcal{F}(x)-F_2(x)]\;\;\;\forall x \in \mathcal{X}\,. \label{eq:p6}
\end{equation}
\textbf{Property 7. (Surjectivity)} The function classifier $\mathcal{F}(x_i)$  is surjective, i.e.\,$\forall \;c_j \in \Omega\;\; \exists x_i \in \mathcal{X}$ such that $F(x_i)=c_j$. In other words $n_j\geq s_j$ and $\sum_{j=1}^k n_j=n$ where the class $c_j$ contains $n_j$ images and $s_j \geq 1$ is the required number of images belonging to $c_j$.

\textbf{Property 8. (Row Dependence)} The $\mathcal{X}$ and $Y$ rows are completely dependent in order to preserve the exactness. Let $\mathcal{X}^\circ$ and $Y^\circ$ be defined above. Then
\begin{equation}
\mathcal{X}^\circ \iff Y^\circ \,.\label{eq:p8}
\end{equation}
\textbf{Property 9. (Useful Information)} Useful information aids in decision making. Redundant information or data which is not useful either causes misclassification or does not influence the result. Let $p=N(\mathcal{X}^c)<N(\mathcal{X})=m$ and $d=m-p$.
\begin{align*}
&F(\mathcal{X}^c,x_i)=y_i \;\;\forall x_i \implies \;\text{$\mathcal{X}^d$ is redundant information, i.e.\,$\mathcal{X}^c$ is a good relevant features subset,}\\
&F(\mathcal{X},x_i)=y_i \;\;\forall x_i \implies \;\text{$\mathcal{X}^d$ is useful information,}\\
&F(\mathcal{X}^c,x_i)\neq y_i \;\text{or}\; F(\mathcal{X},x_i) \neq y_i \;\;\text{for some $x_i$}\;\implies \;\text{we can not decide about $\mathcal{X}^d$.}
\end{align*}
\textbf{Property 10. (Transformations Invariance)} It is desirable that the classifier $F(x_i)$ be invariant under rotation, translation and scaling. For scaling, one can define a degree of scaling in order to avoid misclassification since a too scaled shark can be seen as a simple fish. Let the function $t(x_i)$ denote any transformation of $x_i$. Using the function classifier we expect:
\begin{equation}
F(t(x_i))=\mathcal{F}(x_i) \,. \label{eq:p10}
\end{equation}
\textbf{Property 11. (Classification)} Let $\mu_i(x_q)$ be the probability of $x_q$ to belong to $i^\text{th}$ class where $\mu_i(x_q) \in [0,1]$ and $\sum_i \mu_i(x_q)=1$. The classification is based on the largest probability $\mu_j(x_q)$ to belong to the exact class. 
\begin{equation}
F(x_q)=c_j  \;\;\text{such that}\;\; \mu_j(x_q)=\overset{k}{\underset{i=1}{ \text{max}}}\{ \mu_i(x_q)\} \,. \label{eq:p11}
\end{equation} 
Therefore, if there exist $\mu_i(x_q)$ and $\mu_j(x_q)$ such that $\mu_i(x_q) \approx \mu_j(x_q)$, then we run the risks of False Negative and False Positive.

\textbf{Property 12. (Robustness)} The classifier is robust to noise caused by the weather, wind, camera status and spotter position. Let $x_i^\circ =x_i+n(x_i)$ where $n(x_i)$ denotes the function of noise. Then
\begin{equation}
F(x_i^\circ)=\mathcal{F}(x_i)\,.\label{eq:p12}
\end{equation}
\textbf{Property 13. (Size of Data)} There is no requirement on how $m=N(\mathcal{X})$ and $n=R(\mathcal{X})$ are related. 
\begin{equation}
(n,m) \in \mathbb{N}^2\,. \label{eq:p13}
\end{equation}
\textbf{Property 14. (Binarization)} A good classifier has to detect the pixels forming the foreground $\mathcal{G}$ and the background $\mathcal{B}$. In other words, it should detect boundaries and shapes of the query image $x_q$. With the threshold $\theta$, the intensity function $g(x,y)$ is replaced by the binary function $h(x,y)$ defined: 
\begin{equation}
h(x,y) =
  \begin{cases}
   0 & \text{if } g(x,y) \leq \theta \text{, i.e.}\; g(x,y) \in  \mathcal{B}\,,\\
   1       & \text{otherwise} \text{, i.e.}\; g(x,y) \in  \mathcal{G}\,.
  \end{cases} \label{eq:p14}
\end{equation}

\textbf{Property 15. (Segmentation)} The classifier $F(x_i)$ has to detect and extract different shapes in a given image. Let $x_q$ be a query image and $x_q^j $ be the $j^\text{th}$ shape in $x_q$. Then
\begin{equation}
\sum_j \Gamma[\mathcal{F}(x^j_q)-F(x^j_q)] \leq \Gamma[\mathcal{F}(x_q)-F(x_q))]\;\;\;\forall x^j_q \in x_q \,. \label{eq:p15}
\end{equation}
\textbf{Property 16. (Decision Fusion)} Let $M_i$ be a classifier model, $\delta _i$ be the training error due in $M_i$ and $\delta$ be the error due to the combination of all $M_i$ for $i =\{1,2,\cdots,k\}$ then
\begin{equation}
\delta \leq \overset{k}{\underset{i=1}{\text{min}}}\{\delta _i\}\,. \label{eq:p16}
\end{equation}
This list of properties we consider to be desirable in modelling of the classifier. In what follows, we will focus on the choices of relevant mathematical techniques relative to shark detection and develop a system to satisfy these properties.
\section{Mathematical Analysis}
\begin{figure}[h]
    \centering
        \includegraphics[scale=0.2]{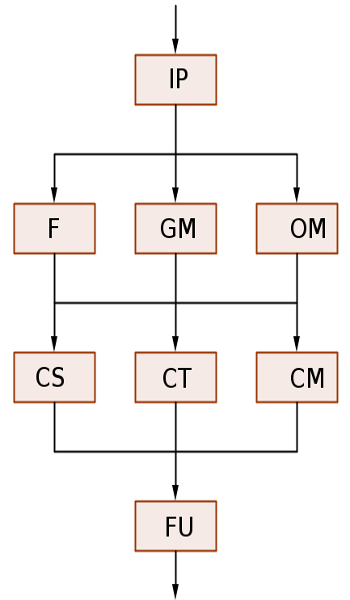}
    \caption{Analysis Result}
    \label{fig:analyse}
\end{figure}

Our analysis based on the characteristics of sharks, the types of cameras used, the climate, wind and so on as above mentioned, and also the techniques that will be used. Before we perform classification, we need to pass through two preliminary steps namely, Image Preprocessing (IP) and Feature Extraction (FE) as shown in Figure \ref{fig:analyse}. Let us look at the role of each step and define the properties required according to our work that is going to lead us to an optimal choice of mathematical technique.

In the first step, Image Preprocessing, we have three tasks: 

\textbf{Filtering}. An image is filtered in order to remove noise and improve contrast. In the case of shark spotting, the noise and bad contrast are due to poor weather conditions

\textbf{Binarization}. An image is binarized in order to detect distinct boundaries. We assume that there are two different sets of information in an image (boundaries and shapes) and we need to assign each pixel to its corresponding set. A good binarization technique has to consider the overall pixel variance in the image. The idea is to look for a threshold $\theta$ and compare each pixel to $\theta$ in order to decide which pixel belongs to the first set and which belongs to the second. 

\textbf{Segmentation}. Segmentation is the process of detecting the different shapes in an image. When we use cameras to spot sharks, we can observe more than one shape. Humans can distinguish different shapes more easily than an automated system can. It will be less expensive computationally to recognize an image having one shape. We need a technique which can segment an image by performing for example, a linear combination between pixels based on their intensities and contrasts. 

In the second step, Feature Extraction, we need to consider only useful information from an image. Nevertheless, with regards to sharks in the ocean, different feature extractors might focus on different types of data, for example space-based or frequency-based data. Shark space-based features due to its movements in space, would be different to its frequency-based features (named F in Figure \ref{fig:analyse}) due to waves produced by its movements. Fortunately, we aim to combine feature extractors to get adversity of features for use in the classification task (see Figure \ref{fig:analyse}). Again a good feature has to be invariant under geometric transformations motivated by previous research in Machine Learning \citep{CMI1}. Then observing different transformations that can be performed on the shark image, we deal with translation, rotation and scaling (TRS). For space-based features, the moment is most used and represents the projection of the image onto the monomial formed by the basis set $\{x, y\}$ \citep{GM}. It is desirable to get a feature extractor that promotes the reconstruction of the image from its set of features. Concerning feature space-based, there are two different moments: one promotes the image reconstruction since its basis is orthogonal (Orthogonal Moment named OM in Figure \ref{fig:analyse}) and the other does not (Geometric Moment named GM).

To assign a query image $x^c_q$ to one of the classes $c_j$ $(j=1,\cdots,k)$, there are several ways to deal with classification: one can evaluate the similarity (CS) between the query image and the rest of the training data followed by votes. Another way would be to reduce the classification error (CM). It defines classification error on the results and seeks to minimize it. Another way would be to train the classifier (CT) from training data so that it can assign a new image to a class using its knowledge. It evaluates training error based on inputs and minimize error. The three approaches are completely different and can produce different results. We need to combine them in order to preserve their individual strengths, i.e.\,get more accurate output since practically, it is not obvious how to choose the best approach. The technique that will be used to combine the classifier outputs (which is called FU in Figure \ref{fig:analyse}) should consider, not only to the outputs obtained by the classifiers, but also to their weaknesses (reliability) and robustness. It should not trust training data at all (uncertainty), i.e.\,it must be a good judge.
\section{Architecture}
Based on the mathematical analysis and classifier properties described in the previous sections, the choices of mathematical techniques are given as follows:

\textbf{Filtering}. We choose Median Filter \citep{PP2} because it replaces each pixel by its nearest neighbours that constructs a more representative set of pixels unlike Average Filter, Highpass Filter or Gaussian Filter. Consequently, the property of Robustness is covered.

\textbf{Binarization}. We use the procedure proposed by Otsu because it is searching the threshold by considering the pixel variances unlike Mean Threshold. This technique covers the property of Binarization.

\textbf{Segmentation}. The Random Walker Technique \citep{PP3} is used instead of Edge detection because it maps an image to a graph whereby the combination of vertices and edges constructs a linear system which is easy to solve. Consequently the property of Segmentation is covered.

\textbf{Feature Extraction.} Considering existing Feature Extractors such as Zernike Moment and Moment. The Complex Moment is chosen as GM, Legendre Moment as OM and Fourier Transform as F relative to Figure \ref{fig:analyse}. Consequently FE covers the properties of Computational Cost, Useful Information and Transformations Invariance. 

\textbf{Classification.} From the analysis made for classification and referring to Figure \ref{fig:analyse}, $k$-Nearest Neighbours is strongly motivated as classifier based on similarity (CS) because of its combination with Genetic Approach. Neural Network is chosen as the classifier based on training (CT) where the error is evaluated by partial derivative (gradient). As a classifier based on misclassification (CM), Support Vector Machine is encouraged to be used since it transforms a classifier to an optimization problem thus ensuring a globally optimal solution. The Classifier covers the properties of Optimality, Graph, Misclassification, Convergence, Uniqueness, Surjectivity, Rows Dependence, Decision and Size of Data.

\textbf{Fusion.} Dempster-Shafer Fusion is chosen over Voting and Bayesian Inference because it combines the classifier outputs based on uncertainty and lack of information. It covers the property on Fusion.
\begin{figure}[h]
    \centering
    \begin{subfigure}[b]{0.30\textwidth}
        \centering
        \includegraphics[width=\textwidth]{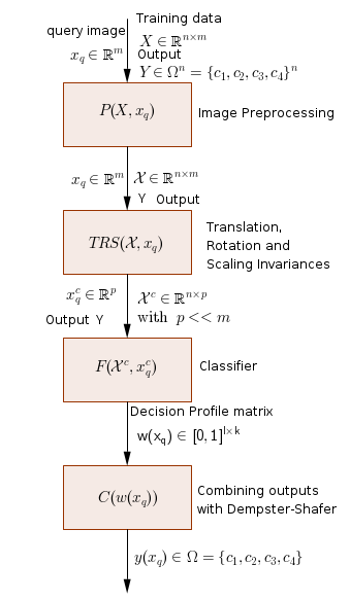}
        \caption{Global Architecture}
        \label{fig:intro1}
    \end{subfigure}
    \begin{subfigure}[b]{0.59\textwidth}
        \centering
        \includegraphics[width=\textwidth]{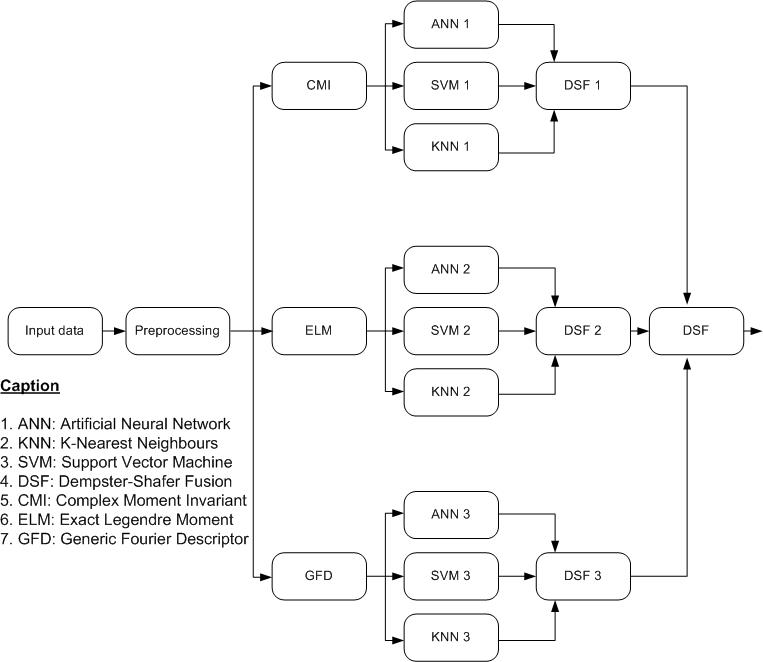}
        \caption{Detailed Architecture}
        \label{fig:intro2}
    \end{subfigure}
    \caption{Shark Detection Architecture}
    \label{fig:intro}
\end{figure}

The architecture \ref{fig:intro2} of our model was built up through a Bottom-Up approach, i.e.\,we have first got an idea of how the architecture had to look and then looked for mathematical techniques that could satisfy the model properties. Figure \ref{fig:intro2} shows the different detailed steps to classify an image and Figure \ref{fig:intro1} shows the global steps by describing the input and output of each step. A detailed explanation of the architecture will be given in the coming chapters. The choice of each technique is motivated also by its computational descriptions.

\chapter{Pretreatments and Transformations}
This chapter will give us a brief detail of Image Preprocessing (IP) and Feature Extraction (FE). IP contains mathematical methods for how a shark image is preprocessed in order to reduce noise caused by wind or weather in the ocean. FE contains mathematical techniques for how a shark image is transformed in order to get relevant features promoting an accurate classification. We start with the raw data $X$ and the query image $x_q$ as inputs and obtain the improved data $\mathcal{X}^c$ and $x_q^c$ as outputs (see Figure \ref{fig:intro}).
\section{Image Preprocessing (IP)}
Let us give the different mathematical techniques chosen for IP and see how the image is transformed in this step.
\subsection{Dimensionality and Filtering}
The first step of IP in our work is filtering, but before that we would like to reduce image size by changing its mode. In the default image mode, each pixel of an image has five-dimensionality $(x,y,i,j,k)$ where $(x,y)$ is the pixel position and $(i,j,k)$ represents code function colour intensities for Red, Green and Blue (RGB) respectively. The new mode (grayscale mode) we need to define has three-dimensionality $(x,y,f)$ where $f$ is the intensity function. We need to map from basis $(x,y,i,j,k)$ to basis $(x,y,f)$ where the function $f(x,y)$ is given as follows:
\begin{equation}
f(x,y)=\dfrac{\alpha i(x,y)+\beta j(x,y)+\gamma k(x,y)}{\mu}\,,\label{eq:IP1}
\end{equation}
where $\alpha, \beta, \gamma \in [0,255]$, $\mu \in \{1,255\}$ and $i,j,k \in [0,255]$. The coefficients $\alpha, \beta, \gamma$ are defined such that $\alpha +\beta +\gamma=1$. Hence $f(x,y) \in [0,1]$ if $\mu=255$ and $f(x,y) \in [0,255]$ if $\mu=1$. Any colour can be obtained from the basis $(i,j,k)$. 
  
\textbf{Median Filter.} After reducing the dimensionality, we now need to remove noise by filtering the image. In Median Filter \citep{PP2}, each pixel is replaced by the median $g(x,y)$ of its nearest neighbour pixels, i.e.\,each pixel is replaced by a representative pixel $g(x,y)$. The number of neighbours to consider is fixed by the user and it is recommended to use an odd number so that the median is exactly one of the existing pixels. As we notice, after applying this filtering, all the nearest neighbours of a pixel have the same grayscale intensity function $g(x,y)$.
\subsection{Binarization and Segmentation}
The second step after reducing noise is to find different shapes in an image (segmentation). We would first like, according to the requirement of the segmentation technique to be used, to get two different sets $\mathcal{G}$ (foreground set) and $\mathcal{B}$ (background set) based on the grayscale intensities $g(x,y)$ of the pixels. In order to get $\mathcal{B}$ and $\mathcal{G}$, some researchers \citep{PP2} used to compare each pixel to the mean pixel $\theta$ considered as a threshold. The mathematical expression of binarization is given by: 
\begin{equation}
h(x,y) =
  \begin{cases}
   0 & \text{if } g(x,y) \leq \theta \text{, i.e.}\; g(x,y) \in  \mathcal{B}\,,\\
   1       & \text{otherwise} \text{, i.e.}\; g(x,y) \in  \mathcal{G}\,,
  \end{cases}
\label{eq:OTSU1}
\end{equation}
where $h(x,y)$ is a binarized function. Unfortunately, this approach does not consider variances in image. Therefore, an optimal approach to find the threshold $\theta$ which considers the pixel variance is given by Otsu \citep{PP2}. Otsu searches for the threshold $\theta$ that minimizes the intraclass variance $\sigma_{in}^2(t)$ and maximizes the interclass variance $\sigma_{out}^2(t)$ in image where $t$ is any threshold candidate. If $t_{\text{min}}=\text{min} (x_q)$, $t_{\text{max}}=\text{max} (x_q)$, $
w_b(t)=\sum_{p=t_\text{min}}^{t-1}\pi(p) \;\text{and } w_a(t)=\sum_{p=t}^{t_\text{max}}\pi(p)\,, $
then
\begin{equation}
\sigma_{in}^2(t)=w_b(t)\sigma_b^2(t)+w_a(t)\sigma_a^2(t) \;\text{and } \sigma_{out}^2(t)=w_a(t)[\mu_b(t)-\mu]^2+w_a(t)[\mu_a(t)-\mu]^2\,, \label{eq:OTSU3}
\end{equation}
where $\sigma_b^2(t)$ and $\sigma_a^2(t)$ are the variances of the pixels in $\mathcal{B}$ and $\mathcal{G}$ respectively, $\mu_b(t)$ and $\mu_a(t)$ are the means of the pixels in $\mathcal{B}$ and $\mathcal{G}$ respectively, $\mu$ is the mean of image pixels and $\pi(p)$ the probability of each pixel $p$ in image $x_q$.

\textbf{Otsu's Algorithm (OA) \citep{PP2}.} The OA is given as follows: (1) Compute probabilities $\pi(p) \;\; \forall p \in x_q$. 
(2) Compute $\sigma_{in}^2(t)$ and $\sigma_{out}^2(t)$ for all thresholds $t$. (3) The optimal threshold $\theta$ corresponds to $t$ which has the maximum of $\sigma^2_{out}(t)$ and the minimum of $\sigma^2_{in}(t)$. In case of more than one optimal threshold, rather their average is considered.

\textbf{Segmentation.} After binarizing the image by detecting the two sets $\mathcal{B}$ and $\mathcal{G}$ using OA as described above, we now need to segment the image $x_q$, i.e.\,to look for different shapes in $x_q$. In image segmentation, each pixel $p_i \in x_q$ is assigned exactly to one shape $x_q^j$ of $x_q$, i.e.\,find the different shapes $x_q^j$ in $\mathcal{G}$ based on probability $\gamma^j_i$ (probability for $p_i$ to belong to $x_q^j$). The algorithm which is going to be used is based on graph theory \citep{PP3} and computes the probability that each vertex of the graph belongs to the different shapes. The image $x_q$ is represented by the graph $G=(P,H)$ where $p_i \in P$ (set of pixels) and  $h_{ij}=(p_i,p_j) \in H$ (set of edges).

The Random Walker Algorithm (RWA) makes the two following assumptions: (1) $G$ is a connected and undirected graph. (2) The neighbours of a pixel are chosen randomly. The motivation for considering RWA as a segmentation technique in our work is due to the fact that it maps randomly an image $x_q$ to a graph $G$. 
Let us see briefly how the graph is weighted and how the segments $x_q^j$ of $x_q$ are found.

The weights of graph $G$ are defined by the Gaussian function $\eta_{ij}$ \citep{PP3}
\begin{equation} 
\eta_{ij} = \exp{\left(-\frac{[g(x_i,y_i) - g(x_j,y_j)]^2}{\sigma}\right) }\,, \label{eq:SEG1}
 \end{equation}
where $g(x_i,y_i)$ is the image intensity at node $p_i$ and $\sigma$  is the standard deviation in $x_q$. The RWA optimizes the following energy \citep{PP3}:
\begin{equation} 
Q(\gamma) = \Upsilon^T M \Upsilon\,, \label{eq:SEG2}
\end{equation} 
where the semi-positive Laplacian matrix $M$ formed by the graph $G$ is given by: 
\begin{equation} M=(\mu_{i,j})_{n \times n},\;\;\;
\mu_{i,j}=
\begin{cases}
d(p_i) & \mbox{if}\ i = j\,, \\
-\eta_{ij} & \mbox{if}\ i \neq j\ \mbox{and}\ p_i \mbox{ is adjacent to } p_j\,, \\
0 & \mbox{otherwise}\,,
\end{cases} \label{eq:SEG3}
\end{equation}
where $\gamma_i$ is the sequence of probabilities of vertex $p_i$ with respect to all shapes $x_q^j$ found in $x_q$ and $d(p_i)$ is the degree of vertex $p_i$. The matrix of probabilities $\Upsilon$ is given by
$
\Upsilon =
 \begin{bmatrix}
  \gamma_1 & \gamma_2 & \cdots & \gamma_n \\
 \end{bmatrix}^{T}\,.\label{eq:SEG4}
$

\textbf{Random Walker Algorithm (RWA).} The RWA is given as follows: (1) Apply Binarization to get the sets $\mathcal{B}$ and $\mathcal{G}$. (2) Map the graph (image) by using Equation (\ref{eq:SEG1}). (3) Solve Equation (\ref{eq:SEG2}). (4) The vertex $p_i$ is assigned to the shape $x_q^k$ such that $\gamma_i^k=\text{max} \{\gamma_i^j\}_{j\geq 1}$. Note that $\sum_j \gamma_i^j=1$.

This leads us to a new step as shown in Figure \ref{fig:intro} whereby a relevant features set of an image, invariant under geometric transformations (scaling, translation and rotation) will be selected.
\section{Feature Extraction (FE)}
As stated above, the aim of a feature extractor is to extract relevant information from an image. Consequently, it reduces image size by considering only useful information. If we are given an image $x_i$ of size $m=r \times q$, then we define the new size $p$ of the image feature such that $p<m$. FE receives the improved training data $\mathcal{X}$ and image query $x_q$ in its input and returns the set of relevant features $\mathcal{X}^c$ of $\mathcal{X}$ and the relevant feature $x_q^c$ of the image $x_q$.

Generally, feature extractors do not provide features which are invariant under all geometric transformations. Therefore, it will be required to normalize feature extractors in order to get features invariant under geometric transformations. Throughout this section, Cartesian coordinates will be used in proving scaling and translation invariance and polar coordinates in proving of rotation invariance. Let us first give a short background of concepts to use in FE as described in Mathematical Analysis.
\subsection{Moments and Fourier Spectra} If the image $t(x_q)$ denotes at least one of the geometric transformations (Translation, Rotation and Scaling TRS) of the image $x_q$, then we aim to look for a set of moment-based features $\mathcal{M}$ and a set of frequency-based features $\mathcal{S}$ such that $\mathcal{M}(t(x_q))=\mathcal{M}(x_q)$ and $\mathcal{S}(t(x_q))=\mathcal{S}(x_q)$. The general formulas of geometric moment \citep{CMI1} and Fourier Spectra \citep{GFD1} are given respectively as follows:
\begin{equation}
\mu_{ab}=\int_{-\infty}^{\infty}\ \int_{-\infty}^{\infty} x^ay^b g(x,y) dx dy\;\text{and } S(u,v)=\int_{-\infty}^{\infty}\ \int_{-\infty}^{\infty} \exp [-2\pi j (u x + v y)] g(x,y) dx dy\,, \label{eq:M1}
\end{equation}
where $a$ and $b$ are moment orders on $x-$axis and $y-$axis respectively and $u$ and $v$ are  the spatial frequencies in vertical and horizontal directions respectively. The centroid $(x_c,y_c)$ for moment on $x$ and $y$ axes of image is given by $x_c=\frac{\mu_{10}}{\mu_{00}}$  and $y_c=\frac{\mu_{01}}{\mu_{00}}$. The centroid $(x_c,y_c)$  for frequency is given by the averages $(\bar{x},\bar{y})$ on $x$ and $y$ axes. 
It follows, by setting new translated variables $x'=x+\kappa$ and $y'=y+\beta$, that
\begin{equation}
x'_c=x_c+\kappa \;\text{and}\; y'_c=y_c+\beta \,. \label{eq:mean1}
\end{equation}
Using (\ref{eq:mean1}), we get $x'-x'_c=x-x_c$ and $y'-y'_c=y-y_c$. This leads us to this following theorem.

\begin{thm}
The central moment and the central Fourier Spectral extract sets of features $\mathcal{M}$ and $\mathcal{S}$ respectively are invariant under translation. \label{thm:central}
\end{thm}

From Equations (\ref{eq:M1}), we can reconstruct the image using their inverse forms. Hence, the pixel intensities obtained from Moment and Fourier Spectra are given respectively by:
\begin{equation}
g(x,y)=\int_{-\infty}^{\infty}\ \int_{-\infty}^{\infty} x^ay^b \mu_{ab} da db\;\;\text{and }\; g(x,y)=\int_{-\infty}^{\infty}\ \int_{-\infty}^{\infty} \exp [-2\pi j (u x + v y)] S(u,v) du dv \,. \label{eq:M2}
\end{equation}
Let us give a useful remark \citep{CMI1} that is going to be used in different proofs in this section.

\begin{rem} If the intensity function $g(x,y)$ has $g'(x',y')$ as its transformed version then $g'(x',y')=g(x,y)\,.$
\end{rem} 

We now introduce the first feature extraction technique based on moments.
\subsection{Complex Moment Invariant (CMI)}
The CMI is the first technique chose. It is an extension of the Moment defined in Equation (\ref{eq:M1}) by including the complex term $j$. The CMI $C_{ab}$ of order $a+b$ in Cartesian and polar coordinates is defined as follows \citep{CMI1}:
\begin{equation}
C_{ab}=\int_{-\infty}^{\infty} \int_{-\infty}^{\infty} (x+j y)^a(x-jy)^b g(x,y) dx dy \;\text{and}\; C_{ab}=\int_{0}^{\infty} \int_{0}^{2\pi} r^{a+b+1}e^{j(a-b)\theta} g(r,\theta) dr d\theta\,, \label{eq:CMI1}
\end{equation}
where $r=\sqrt{x^2+y^2}$, $\theta=\arctan \dfrac{y}{x}$, $y=r \sin \theta$ and $x=r \cos \theta$. We now need to show how CMI extracts a set of features invariant under TRS. 

Firstly, uniform invariance under scaling can be proven by setting $x'=\alpha x$ and $y'=\alpha y$ where $\alpha$ is the scaling factor. Then the new CMI $C'_{ab}$ is:
\begin{align*}
C'_{ab}&=\int_{-\infty}^{\infty} \int_{-\infty}^{\infty} (x'+j y')^a(x'-jy')^b g'(x',y') dx' dy'\\
&=\alpha^{a+b+2} \int_{-\infty}^{\infty} \int_{-\infty}^{\infty}  ( x+j y)^a(x-j y)^b g( x, y) dx dy=\alpha^{a+b+2} C_{ab}\,.
\end{align*}
Since $C'_{ab}=\alpha^{a+b+2} C_{ab}$ is not invariant under scaling, we need to get another moment by normalizing $C_{ab}$. In particular $C'_{00}=\alpha^{2} C_{00}$, hence the new moment is defined as follows:
\begin{align*}
\mathcal{C}'_{ab}=\dfrac{C'_{ab}}{(C'_{00})^w}=\dfrac{\alpha^{a+b+2}C_{ab}}{(\alpha^2C_{00})^w}=\dfrac{\alpha^{a+b+2}}{\alpha^{2w}}\frac{C_{ab}}{(C_{00})^w}=\dfrac{\alpha^{a+b+2}}{\alpha^{2w}} \mathcal{C}_{ab}\,.
\end{align*}
Setting $2w=a+b+2$, we get $\mathcal{C}'_{ab}=\mathcal{C}_{ab}$, i.e.\,the moment $\mathcal{C}_{ab}$ is invariant under scaling with $2w=a+b+2$. 

Secondly, translation invariance is achieved straightforward by the use of Theorem \ref{thm:central}. Finally, the invariance under rotation can be proven by setting $r'= r$ and $\theta'= \theta+\psi$ with $\psi$ the angle of rotation. Then the new complex moment $C'_{ab}$ on $g'( r',  \theta')$ in polar coordinates is
\begin{align*}
C'_{ab}&=\int_{0}^{\infty} \int_{0}^{2\pi} r'^{a+b+1}e^{j(a-b)\theta'} g'(r',\theta') dr' d\theta'=\int_{0}^{\infty} \int_{0}^{2\pi} r^{a+b+1}e^{j(a-b)(\theta+\psi)} g'(r,\theta+\psi) dr d\theta\\
&=\int_{0}^{\infty} \int_{0}^{2\pi} r^{a+b+1}e^{j(a-b)(\theta+\psi)} g(r,\theta) dr d\theta=e^{j(a-b)\psi}C_{ab}\,.
\end{align*}
Since $C_{ab}$ is not invariant under rotation, we need to normalize it. The expression of $C'_{ab}$ leads to the following lemma stated by \cite{CMI1}:

\begin{lem}
Let $g'$ be a rotated function of $g$ such that $g(r,\theta)=g'(r,\theta +\psi)$ where $\psi$ is the angle of rotation. Then
\begin{equation}
C'_{ab}=e^{j(a-b)\psi}C_{ab}\,. \label{eq:CMI3}
\end{equation}
\end{lem}

The expression in (\ref{eq:CMI3}) shows that the moment $C_{ab}$ is invariant under rotation if $a=b$ which is weak since many features such that $a \neq b$ will be lost. We are looking how to normalize this so that the moment is invariant even though $a\neq b$. \cite{CMI1} used a phase cancellation approach by defining $k$ different elements of $a,b$ and $c$ such that $\sum_{j=1}^k c_j(a_j-b_j)=0$. Since $\prod_{i=1}^k e^{j\psi c_i(a_i-b_i)}=e^{j\psi\sum_{i=1}^k  c_i(a_i-b_i)}$, then the normalized moment can be defined as $\mathcal{M}=\prod_{i=1}^k C_{a_ib_i}^{c_i}$. This leads us to modify a theorem (rotation invariance) stated in \citep{CMI1} and get the new theorem (TRS invariance) stated as follows:

\begin{thm}
Let $k\geq 1$, $a_i, b_i$ and $c_i (i=1,\cdots, k)$ be non-negative integers such that $\sum_{i=1}^k c_i(a_i-b_i)=0$. If
\begin{equation}
C_{ab}=\int_{-\infty}^{\infty} \int_{-\infty}^{\infty} \bigg[(x-x_c)+j (y-y_c)\bigg]^a\bigg[(x-x_c)-j(y-y_c)\bigg]^b g(x,y) dx dy \,,\label{eq:CMI6}
\end{equation}
then the moment
\begin{equation}
 \prod_{i=1}^k \mathcal{M}^{ci}_{a_i,b_i}\;\text{where }\; \mathcal{M}^{ci}_{a_i,b_i}=\frac{C_{a_ib_i}^{c_i}}{ C_{00}^{w_i}}\;\text{and }\;  w_i=c_i\frac{a_i+b_i+2}{2}\; \text{is invariant under TRS} \,.\label{eq:CMI7}
\end{equation}
\end{thm}

The new CMI defined in (\ref{eq:CMI7}) extracts the features set $\mathcal{M}$ invariant under TRS. The next step is to construct a set of features by combining the different orders $a_i, b_i$ and $c_i$ of moments. For instance the set of feature $\mathcal{M}=\{\mathcal{M}^{1,2}_{0,2}\mathcal{M}^{1,2}_{2,0}, \mathcal{M}^{2,5}_{1,2}\mathcal{M}^{1,2}_{2,0}, \mathcal{M}^{1,5/2}_{1,2}\mathcal{M}^{1,5/2}_{2,1}\}$ is invariant under TRS. Let us move to the second feature technique based on frequency.
\subsection{Generic Fourier Descriptor (GFD)}
The GFD is based on spatial frequency by transforming the image in the frequency plane using Fourier Transform as stated in Equation (\ref{eq:M1}). The expression of Fourier $F(u,v)$ in polar coordinates \citep{GFD1}:
\begin{equation}
S(\rho,\psi)=\int_{0}^{\infty} \int_{0}^{2\pi}  rg(r,\theta)e^{-2\pi j r \rho \cos (\theta + \psi)} dr d\theta \,,\label{eq:GFD1}
\end{equation}
where $v=\rho \sin \psi$, $u=\rho \cos \psi$, $\theta=\arctan \dfrac{y}{x}$, $r=\sqrt{x^2+y^2}$, $y=r \sin \theta $ and $x=r \cos \theta$. And $(r,\theta)$ are polar coordinates in the image plane and $(\rho, \psi)$ are polar coordinates in frequency space. The bad news is that to normalize the Fourier Transform defined in (\ref{eq:GFD1}) in order to extract features invariant under TRS is not possible \citep{GFD1}. One of the ways to deal with that, as suggested by \cite{GFD1}, is to keep the Fourier Cartesian form $S(u,v)$ in polar form $S(\rho,\psi)$. Therefore, the result in polar coordinates is given as follows:
\begin{equation}
S(\rho,\psi)=\int_{0}^{\infty} \int_{0}^{2\pi}  g(r,\theta)e^{-j2\pi(r \rho+\theta \psi)} dr d\theta \,, \label{eq:GDF4}
\end{equation}
where $\rho$ and $\psi$ are respectively radial frequency and angular frequency. 

To first achieve invariance under uniform scaling, we set the new polar coordinates $r'=\sqrt{x'^2+y'^2}=\sqrt{\alpha^2 x^2+\alpha^2 y^2}=\alpha r$ and $\theta'=\arctan \dfrac{y'}{x'}=\arctan \dfrac{\alpha y}{\alpha x}=\theta$. The new Fourier Transform is given by:
\begin{align*}
S'(\rho,\psi)&=\int_{0}^{\infty} \int_{0}^{2\pi}  g'(r',\theta')e^{-j2\pi(r' \rho+\theta' \psi)} dr' d\theta'=\int_{0}^{\infty} \int_{0}^{2\pi}  g(r,\theta)e^{-j2\pi(\alpha r\rho+\theta \psi)} \alpha dr d\theta\\&=\alpha \int_{0}^{\infty} \int_{0}^{2\pi}  g(r,\theta)e^{-j2\pi(\alpha r\rho+\theta \psi)} dr d\theta \,.
\end{align*}
$S(\rho,\psi)$ is not invariant under scaling. In order to cancel $\alpha$ in $e^{-j2\pi(\alpha r\rho+\theta \psi)}$ of $S'(\rho,\psi)$, we divide the radius $r$ by its maximum value $R=\underset{x,y}{\text{max}}\{r\}$. Therefore Equation (\ref{eq:GDF4}) becomes:
\begin{equation}
S(\rho,\psi)=\int_{0}^{\infty} \int_{0}^{2\pi}  g(r,\theta)e^{-j2\pi(\frac{r}{R} \rho+\theta \psi)} dr d\theta \,.\label{eq:GDF44}
\end{equation}
In order to cancel the factor $\alpha$ in $S'(\rho,\psi)$, since from Equation (\ref{eq:GDF44}) $S'(\rho,\psi)=\alpha S(\rho,\psi)$ and $S'(0,0)=\alpha S(0,0)$, we divide $S(\rho,\psi)$ by $S(0,0)$. Therefore the new Fourier Transform is:
\begin{equation}
\mathcal{S}(\rho,\psi)=\frac{S(\rho,\psi)}{S(0,0)} \,,\label{eq:GDF45}
\end{equation}
where $S(0,0)\neq 0$. It follows by Theorem \ref{thm:central} that the central Fourier Transform $S(\rho,\psi)$ is invariant under translation. It is straightforward that the equation defined in (\ref{eq:GDF45}) is invariant under rotation by setting $r'=r$ and $\theta'=\theta+\phi$. This leads us to the following theorem:
\begin{thm}
Let $r=\sqrt{(x-x_c)^2+(y-y_c)^2}$, $\theta=\arctan \dfrac{y-y_c}{x-x_c}$ and the Fourier Transform $S(\rho,\psi)$ be defined in Equation (\ref{eq:GDF44}). Then the normalized Fourier Transform
\begin{equation}
\mathcal{S}(\rho,\psi)=\frac{S(\rho,\psi)}{S(0,0)} \; \text{is invariant under TRS}\,. \label{eq:GFD46}
\end{equation}
\end{thm}
We do not need to construct systematically as done in CMI the basis which is one of the advantage of Generic Fourier Descriptor. Its set of features looks like:
\begin{equation}
\mathcal{S}=\{\mathcal{S}(0,1),\cdots,\mathcal{S}(1,0),\mathcal{S}(1,1), \cdots\} \in \mathbb{R}^p\,.
\end{equation}
Let us jump to the last Feature Extraction technique based on Legendre moments.
\subsection{Exact Legendre Moment}
The Legendre Moment of order $c=(a+b)$ with the $a^{th}$ and $b^{th}$ order Legendre polynomials $\mathcal{L}_a(x)$ and $\mathcal{L}_b(x)$ is defined as follows \citep{LM2}:
\begin{equation}
M_{ab}=K_{ab}\int_{-1}^1 \int_{-1}^1 \mathcal{L}_a(x)\mathcal{L}_b(y) g(x,y) dx dy \,, \label{eq:LM1}
\end{equation}
where $K_{ab}=\dfrac{(2a+1)(2b+1)}{4}$. According to \cite{LM2}, Rodrigues defined $\mathcal{L}_a(x)$ by
\begin{equation}
\mathcal{L}_a(x)=\frac{1}{2^a a!} \frac{d^a}{dx^a} (x^2-1)^a \;\;a=0,1,2,\cdots \,. \label{eq:LM2}
\end{equation}
It follows from Equation (\ref{eq:LM2}) that $\mathcal{L}_0(x)=1$ and $\mathcal{L}_1(x)=x$. 

Since for $x, y \in [-1,1]$ the set formed by the Legendre Polynomials is completely orthogonal, we need to map the true variables $x$ and $y$ to normalized pixel coordinates $x_i$ and $y_j$ respectively. If the image has size $n\times m$, the normalized pixel coordinates are given by $x_i=-1+i \delta x$ and $y_j=-1+j \delta y$ where  $\delta x =\frac{2}{n}$ and $\delta y =\frac{2}{m}$. It can be proven that $\delta x_i=\delta x$ and $\delta y_j=\delta y$.

The Legendre Polynomial defined in Equation (\ref{eq:LM2}) has recurrence relation defined by 
\begin{equation}
\mathcal{L}_{a+1}(x)=-\dfrac{a\mathcal{L}_{a-1}(x)-(2a+1)x\mathcal{L}_a(x)}{(a+1)}\,.\label{eq:LM4}
\end{equation}
According to \cite{LM2}, because of the computational cost of Legendre moment defined in Equation (\ref{eq:LM1}), its accurate approximation (ELM) is defined as follows:
\begin{equation}
\mathcal{M}_{ab}=\sum_{i=1}^{n} \sum_{j=1}^{m}  I_a(x_i-x_0)I_b(y_j-y_0)g(x_i,y_j)\;\;\text{for}\; a,b \geq 1 \,,\label{eq:LM5}
\end{equation}
where 
$I_a(x_i)=\dfrac{2a+1}{2a+2}\bigg[x_i\mathcal{L}_a(x_i)-\mathcal{L}_{a-1}(x_i) \bigg]^{u_{i+1}}_{u_i}\;\text{and }\;I_b(y_j)=\dfrac{2b+1}{2b+2}\bigg[y_j\mathcal{L}_b(y_j)-\mathcal{L}_{b-1}(y_j) \bigg]^{v_{j+1}}_{v_j} \,.
$
This ELM defined in Equation (\ref{eq:LM5}) is invariant under translation and scaling \citep{LM2} where $x_i,y_j \in [-1,1]$, $(x_o, y_o)$ is the scaled centroid, and the upper and lower limits are defined as follows:
\begin{equation}
u_{i+1}=x_i+\frac{\delta x_i}{2},\; u_{i}=x_i-\frac{\delta x_i}{2}, \; v_{j+1}=y_j+\frac{\delta y_j}{2}  \;\text{and } v_{j}=y_j-\frac{\delta y_j}{2}\,. \label{eq:LM6}
\end{equation}  
So, the set of features obtained from ELM invariant under translation and scaling looks like:
\begin{equation}
\mathcal{M}=\{\mathcal{M}_{11}, \mathcal{M}_{12},\cdots, \mathcal{M}_{21},\mathcal{M}_{22},\cdots\} \in \mathbb{R}^p \,. \label{eq:LM7}
\end{equation} 
To conclude this chapter, Image Preprocessing as a preliminary step has reduced dimensionality of the image, improved the image contrast, reduced noise and detected different shapes in an image. Likely, Feature Extractors extracted features invariant under TRS. As a result, we have got a refined set of training data $\mathcal{X}^c$ and image query $x^c_q$. The $\mathcal{X}^c$ can be either test data used to compute the result accuracy or validation data used to select the best performed model approach.

We introduce the next and last block of techniques which is besides the main block of this work. In this next chapter, three different techniques will be used as classifiers and another technique will be used to fuse classifier outputs in order to improve their outputs.

\chapter{Classification and Fusion Techniques}
As explained in Chapter 2, a classifier has an ability to select a new image (query image) $x_q^c$ from one of the existing classes based on similarity. In this chapter, we will give our understanding of three different classification approaches chosen and the technique used to combine their outputs. 

Each of the classifier models $M_i$ has to assign the image $x_q^c$ in all of the classes in terms of probabilities $\omega^j_{i}(x^c_q)\in [0,1]$.
As inputs we have the training data $\mathcal{X}^c$ and query image $x_q^c$. As outputs we have the decision profile matrix $\omega^j_{i}(x^c_q) \in \omega_{i}(x^c_q)$ where $\omega_{i}(x^c_q)=[\omega^1_i(x_q^c),\cdots,\omega^k_i(x_q^c)]$ as stated in the notation given in Chapter 2. Let us start with the first classification technique.
\section{Artificial Neural Network (ANN)}
ANN is inspired by the human nervous system. The human nervous system is based on \textit{neurons} which process information. 
\begin{figure}[h]
    \centering
        \includegraphics[scale=0.5]{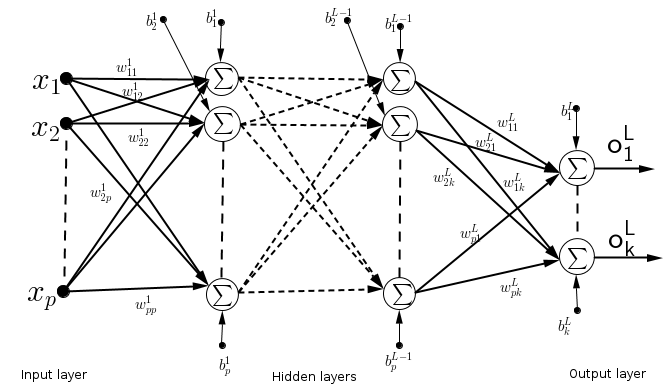}
    \caption{Neural Network Architecture}
    \label{fig:ANN1}
\end{figure}
A neuron receives information sent by another neuron through its \textit{dendrites} and sends the information to other neurons using its \textit{axon}. The biological brain has over $10$ billions neurons and each neuron is connected to the others through $10000$ synapses \citep{ANN1}. ANN guarantees us good learning through its different techniques. The presence of differential equations (gradients), random choice and updates of weights are the keys to ANN success.

Our Neural Network architecture is explained as follows: it contains three layers as shown in Figure \ref{fig:ANN1}. The input layer contains neurons that activate the network, and the only task this layer has is to transmit information. The hidden layer receives information from the input layer and transmits it to the other layers (hidden layers or output layer). The output layer can only receive information from the hidden layer and returns Neural Network output. A network contains exactly one input layer, one output layer and, the number of hidden layers depends on the recognition pattern problem; but it is recommended to use one hidden layer \citep{ANN2}. The ANN architecture can be performed using the One Against All procedure (OAA) as proposed in \cite{ANN3}. There are other procedures such as the P Against Q (PAQ) and the One Against One (OAO). These three procedures show us how layers are connected to one another.

In accordance with the architecture shown in Figure \ref{fig:ANN1},  the goal of ANN is to find the set of weights $w$ that will map accurately to the exact output. Let us now focus on a mathematical description of ANN.
\subsection{MultiLayer Perceptron (MLP)}
According to \cite{ANN2}, it is recommended  to use MLP when we deal with non-linear separable data. MLP architecture contains at least one hidden layer as shown in Figure \ref{fig:ANN1}.

Each neuron, except those of the input layer performs the sum of the information received and actives the sum, i.e.\,it scales the sum to a real value in $[0,1]$. So, our model is defined such that every neuron has the same activation function $g(.)$ and has its own bias $b$. A bias as shown in Figure \ref{fig:ANN1}, helps ANN to shift its output from $0$ to $1$ or vice-versa.

Let $w_{ij}^l$ be the weight of the dendrite from the $i^{th}$ neuron of layer $l-1$ to the $j^{th}$ neuron of layer $l$; $z^l_i$ be the $i^{th}$ neuron of layer $l$; $b^l_i$ be the bias of the $i^{th}$ neuron in the $l^{th}$ layer; and $o^l_i$ be the output of the $i^{th}$ neuron in layer $l$. The idea is to initialize all the weights, compute the output $o^L_j$ in the output layer based on the Feedfoward (FF) approach, i.e.\,the layer $l$ uses the output of layer $l-1$ as its input. At the end of FF, the network output $o^L_j(x_i)$ of neuron $j$ is computed where $x_i$ represents a single image in training data $\mathcal{X}^c$. If the error between the ANN output $o^L_j$ and the exact output $y_j$ is still significant, then update the weights based on Backpropagation (BP) and apply again FF until the error becomes small enough. 

The updates of weights and biases can be developed using two different approaches: \textit{online} and \textit{batch}. 
In this work we choose batch simply because the minimization of the overall error ensures us a good expected result. The \textit{epoch time} is the length of time the classifier learns, i.e.\,the time limit the BP technique is applied. The desired output $y_j$ is a $k$- dimension vector since we have $k$ classes, i.e.\,$y_j=(0,0,\cdots,1)$ if we want to depict the $k^{th}$ class. Let us see how to compute the overall output error which is going to lead us to the description of BP technique.
\subsection{Cost Function}
The cross entropy function \citep{ANN2} is defined by: 
\begin{equation}
E_c(w,b)=-\frac{1}{n}\sum_i \sum_j \bigg[ y_j(x_i) \ln o^L_j(x_i) +(1-y_j(x_i)) \ln (1-o^L_j(x_i))\bigg]\,, \label{eq:entropy}
\end{equation}
where $o_j(x_i)=g(z_j)=g\bigg( \sum_k w_{jk} x_j+b_j \bigg)$ and  $g(.)$ the sigmoid activation function given by $g(z)=[1+\exp ({-z})]^{-1}$. It follows that:
\begin{equation}
g'(z)=g(z)[1-g(z)] \,. \label{eq:c7}
\end{equation}
The entropy has the advantage that as it gets close to zero when the neuron's output becomes close to the exact output and the larger the error, the faster the classifier (unlike quadratic cost function for instance). The changes of $w$ and $b$ are determined by using the Backpropagation (BP) approach. BP introduces the intermediate quantity $\delta_i^l$ representing the error in the $i^{th}$ neuron of layer $l$. Let us now perform mathematical analysis of BP by developing the differential equations of the cost error $E_c(w,b)$ defined in \ref{eq:entropy} with respect to weights and biases.
\subsection{Backpropagation Technique (BP)}
The BP gives the procedure to compute $\delta_i^l$. The soma $z^l_j$ and the output $o^l_j$ are given by:
\begin{equation}
z^l_j=\sum_k w_{jk}^l o^{l-1}_k +b_j^l \;\text{and } o^l_j=g(z^l_j)\,.\label{eq:z}
\end{equation}
Let us perform the partial derivative of $E_c$ with respect to weights $w^L_{jk}$ and bias $b^L_j$ where $L$ represents the ANN output layer. For ergonomic reason, the outputs $y_j(x_i)$ and $o^{L}_k(x_i)$ will be expressed by $y_j$ and $o^{L}_k$ respectively. Hence,
\begin{align*}
\frac{\partial E_c}{ \partial w^L_{jk}}&= -\frac{1}{n}\sum_i  \bigg( y_j \frac{o^{L-1}_k g'(z^L_j)}{g(z^L_j)} -(1-y_j)\frac{o_k^{L-1} g'(z^L_j)}{1-g(z^L_j)}\bigg)\,,\\
&= -\frac{1}{n}\sum_i  \bigg(  \frac{ y_j-g(z^L_j)}{g(z^L_j)(1-g(z^L_j))} \bigg)o_k^{L-1}g'(z^L_j)\,.
\end{align*}
Using Equations (\ref{eq:c7}) and (\ref{eq:z}), 
\begin{equation}
\frac{\partial E_c}{ \partial w^L_{jk}}=\frac{1}{n}\sum_i  \bigg(   o^L_j-y_j \bigg)o_k^{L-1}\,. \label{eq:c8}
\end{equation}
\begin{align*}
\frac{\partial E_c}{ \partial b^L_{j}}&=  -\frac{1}{n}\sum_i  \bigg( y_j \frac{ g'(z^L_j)}{g(z^L_j)} -(1-y_j)\frac{ g'(z^L_j)}{1-g(z^L_j)}\bigg)\,,\\
&=  -\frac{1}{n}\sum_i  \bigg(  \frac{ y_j-g(z^L_j)}{g(z^L_j)(1-g(z^L_j))} \bigg)g'(z^L_j)\,. \\
\end{align*}
Using Equations (\ref{eq:c7}) and (\ref{eq:z}), 
\begin{equation}
\frac{\partial E_c}{ \partial b^L_{j}}=\frac{1}{n}\sum_i  \bigg(o^L_j-y_j \bigg)\,. \label{eq:c9}
\end{equation}
Let us notice that in each output neuron, there is an error due to computation, i.e.\,during the computation of the ANN output $o^l_j$, the neuron instead of performing $o^l_j=g(z^l_j)$, it performs $y^l_j=g(z^l_j+\Delta z^l_j)$. One can say that this is good news, since the error improves the way the network learns. \cite{ANN2} defined the error $\delta^l_j$ such as:
\begin{equation}
\frac{\partial E_c}{\partial w^l_{jk}}=\frac{\partial E_c}{\partial z^l_j} \frac{\partial z^l_j}{\partial w^l_{jk}}\;\text{ where  } \delta^l_j=\frac{\partial E_c}{\partial z^l_j}\,.\label{eq:c10}
\end{equation}
Using the definition of $z_j^l$ stated in Equation (\ref{eq:z}), $\frac{\partial z^l_j}{\partial w^l_{jk}}=o_k^{l-1}$. Let us now compute the neuron error $\delta_j^L$ in the output layer $L$.
\begin{equation}
\delta_j^L=\frac{\partial E_c}{ \partial z^l_j}
=  -\frac{1}{n}\sum_i  \bigg(  \frac{ y_j-g(z^L_j)}{g(z^L_j)(1-g(z^L_j))} \bigg)g'(z^L_j)=\frac{1}{n}\sum_i  \bigg(  g(z^L_j)-y_j \bigg)=\frac{1}{n}\sum_i \bigg(  o^L_j-y_j \bigg)\,.\label{eq:erro}
\end{equation}
So, the expressions in Equations (\ref{eq:c8}) and (\ref{eq:c9}) can be expressed in terms of $\delta^L_j$ by:
\begin{equation}
\frac{\partial E_c}{\partial w^L_{jk}}=\delta^L_j o^{L-1}_k \;\;\text{and } \frac{\partial E_c}{\partial b^L_{j}}=\delta^L_j\,. \label{eq:c11}
\end{equation}
Since we have $\delta_j^L$, we need to see if we can express the errors $\delta_j^l$ in the hidden layers. It is better to start with the layer $L-1$ which is close to the output and then generalize it for any hidden layer $l$. 
\begin{align*}
\frac{\partial E_c}{ \partial w^{L-1}_{ij}}
&=  \frac{1}{n} \sum_k  \sum_i  \bigg(  g(z^L_k)-y_k\bigg)\frac{\partial z^L_k}{ \partial w^{L-1}_{ij}}= \frac{1}{n} \sum_k  \sum_i  \bigg(  g(z^L_k)-y_k\bigg)\frac{\partial z^L_k}{ \partial o^{L-1}_{j}} \frac{\partial o^{L-1}_j}{ \partial w^{L-1}_{ij}}\,.\\
\end{align*}
Using Equation (\ref{eq:z}) in the above equation, we have
\begin{align*}
\frac{\partial E_c}{ \partial w^{L-1}_{ij}}&=\frac{1}{n} \sum_k  \sum_i  \bigg(  g(z^L_k)-y_k\bigg)w^L_{jk}\frac{\partial o^{L-1}_j}{ \partial w^{L-1}_{ij}}= \frac{\partial o^{L-1}_j}{ \partial w^{L-1}_{ij}}\ \bigg[ \sum_k w^L_{jk} \bigg( \frac{1}{n} \sum_i  \bigg(  g(z^L_k)-y_k\bigg) \bigg) \bigg]\,,\\
&= \frac{\partial o^{L-1}_j}{ \partial w^{L-1}_{ij}}\  \sum_k w^L_{jk}\delta^L_k =g'(z^{L-1}_j) \frac{\partial z^{L-1}_j}{ \partial w^{L-1}_{ij}}  \sum_k w^L_{jk}\delta^L_k=g(z^{L-1}_j)[1-g(z^{L-1}_j)] o^{L-2}_j  \sum_k w^L_{jk}\delta^L_k \,. 
\end{align*}
Finally, \begin{equation}
\frac{\partial E_c}{ \partial w^{L-1}_{ij}}=o^{L-2}_j \delta^{L-1}_j\;\;\text{where } \delta^{L-1}_j= g(z^{L-1}_j)(1-g(z^{L-1}_j))  \sum_k w^L_{jk}\delta^L_k\,.
\end{equation}
Since we expressed $\delta^{L-1}_j$ in terms of $\delta^{L}_i$, we can thus write the error $\delta^{l}_j$ in terms of $\delta^{l+1}_i$ for all neuron $i$ of layer $l+1$. More generally the error in layer $l$ is given by:
\begin{equation}
\delta^{l}_j= o^{l+1}_j[1-o^{l+1}_j]  \sum_i w^{l+1}_{ji}\delta^{l+1}_i\,. \label{eq:back}
\end{equation}
As we have seen how the weights and biases change throughout the process using BP, the next step is to show how the weights and biases are updated.
\subsection{Update of weights and biases}
Stochastic Gradient Descent (SGD) is used to update the weights and biases. SGD is chosen because of the random choice of weights and biases which avoids converging with the local minimal of $E_c$ and the consideration of the gradients. Hence, the new weights and biases are given respectively by \cite{SGD}:
\begin{equation}
w^l_{ij}\to w^l_{ij}-\beta\sum_j o^{l-1}_j \delta^l_j\;\;\text{and  } b^l_{j}\to b^l_{j}-\beta\sum_j  \delta^l_j \,, \label{eq:sgd1}
\end{equation}
with the learning rate $\beta$ chosen adequately. The expressions $o^{l-1}_j \delta^l_j$ and $\delta^l_j$ are respectively the gradients of $E_c(w,b)$ over $w^l_{ij}$ and $b^l_{j}$. It is shown in \citep{SGD} that SGD provides an optimal convergence of the weights and biases.
\subsection{ANN Algorithm}
In summary, the number of hidden layers, the length of epoch $T$ and the learning rate $\beta$ define the ANN architecture. Thus there is no exact theoretical way to design ANN architecture, except by experiment. By experience and through some papers read \citep{ANN2}, one hidden layer is recommended since more than one hidden layer improves the risk of convergence to a local minimal. So, the ANN can be summarised as follows:

(1) Initialize randomly the weights and biases in ANN in order to avoid a local minimum. (2) Apply the FF approach to get Neural output using Equation (\ref{eq:z}). (3) Compute the output error $\delta ^L_j $ using Equation (\ref{eq:erro}) and apply BP to compute the output error in the rest of the hidden layers using Equation (\ref{eq:back}). (4) Apply SGD to update the weights and biases using Equation (\ref{eq:sgd1}). Let us now give a toy example in order to improve our understanding of ANN.

\begin{exa}
We are given two inputs neurons with an image $x=[1,\; 0.5]$, one output neuron with $y=0$ the exact output and the learning rate $\beta=0.1$. (1) We initialized the weights to $w_1=0.5$ and $w_2=0.4$, and the bias to $b=0.7$. (2) $z=\sum_i w_i x_i +b=0.5 \times 1 + 0.4 \times 0.5 + 0.7=1.4$. The ANN output is  $o=g(z)=0.802$. (3) The error in the output is $\delta=0.802-0=0.802$. (4) The new weights are $w_1=w_1-\beta x_1 \delta=0.5-0.1 \times 1 \times 0.802=0.4198$ and $w_2=w_2-\beta x_2 \delta=0.4-0.1 \times 0.5 \times 0.802=0.3599$, the new bias is $b=b-\beta  \delta=0.7-0.1 \times 0.802=0.6198$. For this first step, the entropy error is $E_c(w,b)=1.619488$. Apply FF again until the error $E_c$ becomes close to zero.
\end{exa}
Let us jump to the second classification technique chosen. This new technique is completely different to ANN but encouraged through its different approaches that we will see very soon. 
\section{$k$-Nearest Neighbours (KNN)}
The KNN is the second classification technique chosen in this work. KNN looks for $k$ images that are very close to the query image based on similarity (fitness). A query image $x_q$ is assigned to class $c_k$ having more images among $k$ images chosen \citep{KNN1}. That means we look for $k$ images of set $\mathcal{X}_k$ which dominate the rest of the images in $\mathcal{X} \setminus \mathcal{X}_k$. If $x_k\in \mathcal{X}_k$ then $x_k \succ x_i\;\;\;\forall x_i \in \mathcal{X}\setminus \mathcal{X}_k$ where $\succ$ expresses the closeness between query image $x_q$ and any image in $\mathcal{X}$.

Dealing with a query image $x_q$ and training data $\mathcal{X}$, the KNN first computes the fitness $f_{x_q}(j)$ between $x_q$ and all the images $x_j \in \mathcal{X}$ which is given by the Mahalanobis distance \citep{MAHA}: 
\begin{equation}
f_{x_q}(j)= \sqrt{(x_q-x_j)^t S^{-1} (x_q-x_j)}\,. \label{eq:MAHA}
\end{equation}
The Mahalanobis distance is preferred in this case because of the consideration of covariances $S$ into account which removes all linear statistical dependencies from data $\mathcal{X}$. To compute the fitness of all images in training data shows the weakness of KNN. Therefore, the following subsections show how KNN is improved by combining it with Genetic Approach (GA). One says that GA improves KNN since images in training data are no longer all considered.
\subsection{GA Terminology}
GA initializes the population $P_0\in \mathcal{X}$ and in each step, it improves the population based on genes without increasing its size. Each image $x_i \in \mathcal{X}$ has a genetic representation called \textit{chromosome} $\kappa_i$. An image $x_i$ is a sequence of real numbers as described in Chapter 2. Then the $25^{\text{th}}$ image in $\mathcal{X}$ looks like $[x_{25}^1,x_{25}^2,\cdots, x_{25}^p]=[0.1,0.253,\cdots, 0.634]$; so its corresponding chromosome is  $\kappa_{25}=\underbrace{00\cdots 25}_p$ where $p$ is the number of \textit{genes} in chromosome $\kappa_i$. GA is based on two main operators: \textit{crossover} and \textit{mutation}. \textit{Crossover} $\Lambda$ creates a new chromosome (offspring $\varrho$) from an existing chromosome (parent $\rho$); the offspring obtained from two parents has to inherit their good genes ($\Lambda(\rho_1, \rho_2) \implies \varrho$). \textit{Mutation} $\lambda$ changes the characteristics of the chromosomes a little such that, if $\kappa_{\text{new}} =\lambda(\kappa_i)$ then $\kappa_{\text{new}} \approx \kappa_i$. So far we have a general understanding of GA; let us show how each of the operators stated above works.
\subsection{Genetic Operators \citep{GA1}}
Let us assume that each chromosome has length $r$ in binary representation.

\textbf{Crossover $\Lambda$.} In population $P_t$, we select two parents $\rho_1$ and $\rho_2$, generate a crossover point $v$ randomly and create two binary masks $m_1=2^r-2^v$ and $m_2=2^r-1$. The two offspring are computed as follows: $\varrho_1=(\rho_1 \;\land \; m_1) \;\lor \; (\rho_2 \; \land \; m_2)$ and $\varrho_2=(\rho_2 \;\land\; m_1) \;\lor\; (\rho_1 \;\land \; m_2)$. All the offspring generated $\varrho_i$ belong to the offspring set $Q_t$. 

\textbf{Mutation $\lambda$.} In population $T_t=Q_t \cup P_t$, we select a chromosome $\kappa_i$, generate a gene $v$ randomly and create a binary mask $m=2^v$. The new chromosome $\kappa_{new}=\kappa_i \;\;\text{ XOR }\;\; m$. The selection of chromosomes of new generation $P_{t+1}$ is based completely on their fitness function $f_{x_q}(i)$ defined in Equation (\ref{eq:MAHA}). Roulette Wheel Selection \citep{GA1} is one of the reproduction techniques used to generate $P_{t+1}$. Since mutation and crossover operators are random, then the $k$ images selected by GA are dominant.
\subsection{GKNN Algorithm}
In summary, the GKNN is given as follows: (1) Generate randomly the first population $P_0$ of $k$ images and evaluate their fitness. (2) Apply the crossover operator to generate offspring set $Q_t$. (3) Change the characteristics of each element (mutation) $\kappa_i \in T_t=P_t \cup Q_t$ and generate new set $N_t$. (4) Select $k$ images from $N_t=T_t \cup P_t$ based on their fitness and assign them to the new population $P_{t+1}$. The stopping criterion occurs when the highest fitness in the current population is less than or equal to that of the previous population. (5) The class of query image $x_q$ corresponds to the class of image having more images among $k$ images of $P_{t+1}$ (voting). The vote can involve probabilities \citep{KNN1}. Hence, the class of query image $x_q$ corresponds to the class of image having the largest probability.

\begin{exa}
We are given the training data $\mathcal{X}=\{(1,1), (0,1), (2,3), (2,2), (1,1), (4,2)\}$, the sequence of outputs $Y=[1, 1, 2, 2, 1, 2]$ and the set of chromosomes $\kappa=\{001, 010, 011, 100, 101, 110\}$. We need to classify the query object $x=(1,\; 0)$. (1) We initialize $P_0=\{x_1, x_2\}$. Using the fitness function defined in Equation (\ref{eq:MAHA}), we have $f_x(1)=1.53$ and $f_x(2)=2.198$. (2) Apply crossover in $P_0$ with $v=2$ and $r=3$. Then $m_1=100$ and $m_2=111$. The offspring are $\varrho_1=(001 \;\land \; 100) \;\lor \; (010 \; \land \; 111)=010$ and $\varrho_2=(001 \;\land \; 111) \;\lor \; (010 \; \land \; 100)=001$. Therefore, $T_0=\{\varrho_1, \varrho_2\}$. (3) Mutate $T_0$ elements based on their genes. So, with $v=2$ we have $m=100$. Therefore, $\kappa_1^{\text{new}}=001 \text{ XOR }100=101=\kappa_5 \implies x_5$, $\kappa_2^{\text{new}}=010 \text{ XOR }100=110=\kappa_6 \implies x_6$. So the set $N_0=\{x_1, x_2, x_5, x_6\}$. (4) For generating the new generation $P_1$, we need the fitness of $x_5$ and $x_6$ which are $f_x(5)=1.53$ and $f_x(6)=2.614$. Since $f_x(1)=f_x(5)<f_x(2)<f_x(6)$, then $P_1=\{x_1, x_5\}$. As we can notice that $P_0 \neq P_1$, but after applying the same procedure in $P_1$, we will get $P_2=\{x_1, x_5\}=P_1$. Therefore, we vote for $x \in c_1$ since $x_1$ and $x_5 \in c_1$.
\end{exa}

With GKNN the computational cost is reduced and the $k$ images selected are more similar because of the type of fitness function chosen and the random change of genes during the selection process. Let us now introduce the last classification technique. 
\section{Support Vector Machine (SVM)}
\begin{figure}[h]
    \centering
    \begin{subfigure}[b]{0.22\textwidth}
        \centering
        \includegraphics[width=\textwidth]{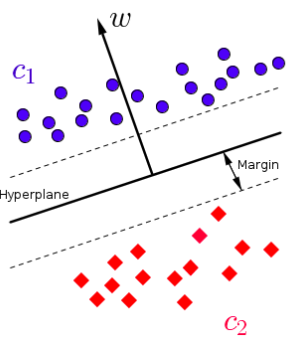}
        \caption{Binary classification}
        \label{fig:svm1}
    \end{subfigure}
    \hfill
    \begin{subfigure}[b]{0.20\textwidth}
        \centering
        \includegraphics[width=\textwidth]{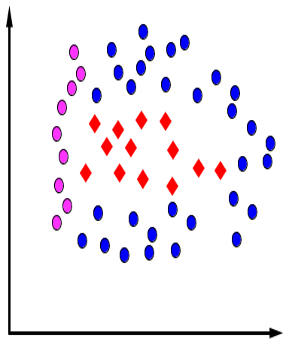}
        \caption{NL separable}
        \label{fig:svm2}
    \end{subfigure}
    \hfill
    \begin{subfigure}[b]{0.22\textwidth}
        \centering
        \includegraphics[width=\textwidth]{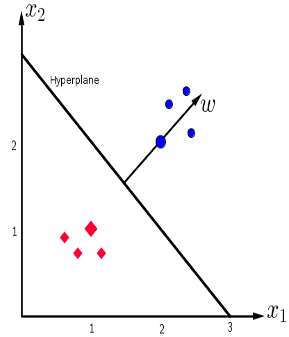}
        \caption{Toy example}
        \label{fig:svm3}
    \end{subfigure}
    \caption{SVM}
    \label{fig:svm}
\end{figure}

SVM is one of the most used classification techniques. Its functionality extends that of ANN. SVM transforms a non-linear problem in the original space to a linear problem in the feature space by using the kernel function. 
 SVM classifies data by increasing the margin between data classes while keeping data separable \citep{ANN1} as shown in Figure \ref{fig:svm1}. Following in \cite{SVM1}, we call support vectors (SV), the points nearest to the separating hyperplane $h$. 

It follows that, dealing with linear separable data, SVM easily achieves the classification task. The hyperplane is of the form:
\begin{equation}
h= \{x  | <w,x> + b =0\}\,. \label{eq:svm1}
\end{equation}
Let us now describe a binary classification with linear separable data in the following subsection in order to get a good understanding of SVM.
\subsection{Binary classification (BC)}
We call BC any mathematical problem which has two different outcomes as shown in Figure \ref{fig:svm1}. The points in the direction of $w$ belong to class $c_1$ and in the opposite direction belong to class $c_2$. The variable $b$ is an offset allowing the displacement of the hyperplane and the normal vector $w$ determines the orientation of the hyperplane. From a mathematical point of view, $w$ is the weighted sum of the support vectors (SV). SVM can detect more than one hyperplane and decides which one is the best. The classifier is given by 
\begin{equation}
F(x)=w \cdot x^T+b \,. \label{eq:svm3}
\end{equation}
The decision is taken such that for all $x_i \in c_1$, $F(x_i)\geq 1$ and for all $x_i \in c_2$, $F(x_i)\leq -1$. Since we know how BC works, let us introduce a simple toy example in order to give more understanding of BC.

\begin{exa}
From Figure \ref{fig:svm3}, the hyperplane is well placed between the two points $x_1=(1,1)$ and $x_2=(2,2)$. Clearly the support vector is $w=(2,2)-(1,1)=(1,1)$. More generally, we can write $w=(a,a)$. Using Equation (\ref{eq:svm3}), we have $F(x_1)=-1 \implies 2a+b=-1$ and $F(x_2)=1 \implies 4a+b=1$. The solution of these two equations is given by $a=1$ and $b=-3$. Hence, $F(x)=x_1+x_2-3$. The decision can be shown as follows, with $x=(1,0)$. For instance, $F(x)=-2 \leq -1 \implies (1,0) \in c_1$. In this example, the SV is directly determined but generally that is not the case. Therefore, a classification task is converted to an optimization problem whereby we seek to minimize the length of $w$.
\end{exa}

Based on the hypothesis saying that data is non-linear (NL) separable, it is difficult to get a boundary between classes in original space. The idea is to apply Kernel function in order to transform this problem to a linear separable one in feature space using inner product. Let us jump to the multiclassification problem with non linear separable data which is our case.
\subsection{Multiclassification}
Let $\mathcal{X}$, $x_i$ and $y_i$ be as defined above. According to \cite{SVM1}, the classifier on $\mathcal{X}$ is defined as follows:
\begin{equation}
F_S(x_q)=\text{arg } \overset{k}{\underset{j=1}{\text{max}}} \{  \langle S_{c_j}, x_q \rangle \} \;\text{with }\; S=\begin{pmatrix}
S_{c_1}\\S_{c_2}\\\vdots\\S_{c_k}
\end{pmatrix}\;\;\text{and }\;\; S_{c_j} \in \mathbb{R}^p\,.\label{eq: svm4}
\end{equation}
The inner product $\langle S_{c_j}, x_q \rangle$ represents the confidence for image $x_q$ to belong to the $j^{th}$ class. 
 Based on the property of Misclassification defined in Chapter 2, the empirical error \citep{SVM1} is defined:
\begin{equation}
\xi_\mathcal{X}(S)=\frac{1}{n} \sum_{i=1}^n \Pi \bigg( F_S(x_i) \neq y_i \bigg)\;\text{where }\; \Pi(\lambda)=\begin{cases}1\;\;\text{if } \lambda \text{ is true}\,,\\
0\;\;\;\text{otherwise}\,.
\end{cases} \label{eq: svm6}
\end{equation}
In \cite{SVM1}, the error $\xi_\mathcal{X}(S)$ of a single example $(x_i,y_i)$ written in Equation (\ref{eq: svm6}) can be rewritten and bounded as:
\begin{equation}
\Pi \bigg( F_S(x_i) \neq y_i \bigg)\leq \overset{k}{\underset{j=1}{\text{max}}} \bigg\{ S_{c_j} x_i +1 -\delta_{y_ic_j}\bigg\}-S_{y_i} x_i \;\; \text{where }\;\; \delta_{y_ic_j}=\begin{cases}1\;\;\text{if } y_i={c_j}\,,\\
0\;\;\;\text{otherwise}\,.
\end{cases} \label{eq: svm7}
\end{equation}
It follows from Equations (\ref{eq: svm6}) and (\ref{eq: svm7}) that:
\begin{equation}
\xi_\mathcal{X}(S)\leq \frac{1}{n} \sum_{i=1}^n  \bigg[ \overset{k}{\underset{j=1}{\text{max}}} \bigg\{ S_{c_j} x_i +1 -\delta_{y_i{c_j}}\bigg\}-S_{y_i} x_i \bigg]\,. \label{eq: svm8}
\end{equation}
It is shown in \cite{SVM1} that 
$\overset{k}{\underset{j=1}{\text{max}}} \{ S_{c_j} x_i +1 -\delta_{y_i{c_j}}\}-S_{y_i} x_i=\varepsilon_i\;\;\;\forall i$ if $\mathcal{X}$ is not linear separable with the slack variable $\varepsilon_i \geq 0$. It follows that:
\begin{equation}
\overset{k}{\underset{j=1}{\text{max}}} \{ S_{c_j} x_i +1 -\delta_{y_i{c_j}}\}-S_{y_i} x_i=\varepsilon_i \implies  S_{c_j} x_i +1 -\delta_{y_i{c_j}}-S_{y_i} x_i \leq \varepsilon_i \,.\label{eq: svm9}
\end{equation}
We need to look for a matrix  $S$ such that $\xi_\mathcal{X}(S)$ is minimized as much as possible. So, we can transform this matter to an optimization problem and use Karush-Kuhn-Tucker (KKT) conditions \citep{OPT1} to solve it.
\subsection{Optimization problem}
We need to apply KKT to solve this optimization problem, thus we need to deal with a convex problem which is always solvable. Minimizing the matrix $S$ subject to constraint defined in Equation (\ref{eq: svm9}) can be formulated as follows where $A$ is the regularization constant: 
\begin{equation}
\begin{aligned}
& \underset{S,\varepsilon}{\text{minimize}}
& & A \|S\|^2_2+\sum_{i=1}^n \varepsilon_i\,, \\
& \text{subject to:}
& & S_{c_j} x_i-S_{y_i} x_i-\delta_{y_i{c_j}} - \varepsilon_i+1 \leq 0\;\;\;\forall i,j \,.
\end{aligned} \label{eq:opt1}
\end{equation}
The next step is to get the dual form of equation \ref{eq:opt1} in order to consider high space dimension, use the Quadratic Programming Technique and use the Kernel function of the form $\kappa(x_1,x_2)=\langle x_1,x_2 \rangle$. The KKT theorem defined in \cite{OPT1} leads to the Lagrangian $L(S,\varepsilon,\lambda )$ where $\lambda$ is the dual variable, i.e.\,$\text{min }\; L(S,\varepsilon,\lambda ) \implies \text{max }\; D(\lambda)$. So the primal problem defined in Equation (\ref{eq:opt1}) can be written in Lagrangian form as follows:
\begin{equation}
 L(S,\varepsilon,\lambda )=A \sum_{j=1}^k \|S_{c_j}\|^2_2+\sum_{i=1}^n \varepsilon_i+\sum_{i=1}^n\sum_{j=1}^k \lambda_{i{c_j}} \bigg[ S_{c_j} x_i-S_{y_i} x_i-\delta_{y_i{c_j}}  -\varepsilon_i+1 \bigg]\;\;\text{subject to } \lambda_{i{c_j}}\geq 0\;\; \; \forall i,j\,.\label{eq:opt2}
\end{equation}
Since the variables $S$ and $\varepsilon$ are primal, the minimality of primal problem $L(S,\varepsilon,\lambda )$ involves that $\dfrac{\partial L}{\partial S_{c_j}}=0$ and $\dfrac{\partial L}{\partial \varepsilon_i}=0$. Therefore \citep{SVM1},
\begin{equation}
\dfrac{\partial L}{\partial \varepsilon_i}=0 \implies \sum_{j}^k \lambda_{i{c_j}}=1 \;\text{and}\; \dfrac{\partial L}{\partial S_{c_j}}=0 \implies  S_{c_j}=A^{-1} \bigg[  \sum_{i=1}^n \bigg(\delta_{y_i c_j}-\lambda_{ic_j}\bigg) \cdot x_i\bigg]\,.\label{eq:conf}
\end{equation}
It is proven \citep{SVM1} that the dual $\text{max }\; D(\lambda)$ of the primal $\text{min }\; L(S,\varepsilon,\lambda )$ which is given by
\begin{equation}
\begin{aligned}
& \underset{\lambda}{\text{maximize}}
& & A \sum_{i=1}^n \eta_i \cdot \delta_{y_i} - \sum_{i=1}^n\sum_{j=1}^k \kappa(x_i, x_j) (\eta_i \cdot \eta_j)\,, \\
& \text{subject to:}
& & \eta_i\leq \delta_{y_i} \;\text{and }\; \eta_i \delta=0\;\;\;\forall i \,,
\end{aligned} \label{eq:opt3}
\end{equation}
its objective function $D(\lambda)$ is concave. Hence, it admits an unique solution where $\eta_i=\delta_{y_i}-\lambda_i=[\eta_{ic_1},\cdots,\eta_{ic_k}]=[\delta_{y_ic_1}-\lambda_{ic_1},\cdots,\delta_{y_ic_k}-\lambda_{ic_k}]$, $\lambda_i=[\lambda_{ic_1}, \cdots, \lambda_{ic_k}]$, $\delta_{y_i}=[\delta_{y_ic_1}, \cdots, \lambda_{y_ic_k}]$, $\delta=\{1\}^k=[1,\cdots, 1]$ and the kernel $\kappa(x_i,x_j)=x_i \cdot x_j$. Using the new notation above, the confidence vector of class $c_j$ defined in Equation (\ref{eq:conf}) can be rewritten as follows:
\begin{equation}
S_{c_j}=A^{-1}  \sum_{i=1}^n \eta_{ic_j} \cdot x_i\,.
\end{equation}
This leads us to the new formulation of the classifier defined in Equation (\ref{eq: svm4}). The new classifier \citep{SVM1} is given by:
\begin{equation}
F_S(x_q)=\text{arg } \overset{k}{\underset{j=1}{\text{max}}} \bigg \{  \sum_{i=1}^n \eta_{ic_j } \cdot \kappa(x_i, x_q) \bigg \} \,. \label{eq: svm40}
\end{equation}
\subsection{SVM Algorithm} In summary, the SVM is given as follows: (1) Solve the optimization problem stated in (\ref{eq:opt3}) using the Karush-Kruhn Tucker theorem. (2) Compute $\sum_{i=1}^n \eta_{ic_j} \cdot \kappa(x_i, x_q)$ for $j=1, \cdots, k$. (3) Finally the class of the query image $x_q$ corresponds to $c_l$ such that $\eta_{ic_l}=\text{arg } \overset{k}{\underset{j=1}{\text{max}}} \{  \sum_{i=1}^n \eta_{ic_j} \cdot \kappa(x_i, x_q)  \}$.

It is shown in Figure \ref{fig:intro} that three different feature extraction techniques (FE) are used. Applying each of the three classification techniques on every FE, we obtain a model with nine classifiers $M_i$. Each model $M_i$ at this stage returns the $i^{\text{th}}$ row of decision matrix profile $w(x^c_q)$. Therefore, this following technique combines the different classifier outputs in order to obtain an unique improved output.
\section{Dempster-Shafer Fusion (DSF)}
The main and only motivation of combining classification outputs is to reduce error in order to solve property Decision Fusion. The choice of DSF is because of the uncertainties in decision profile matrix described above is solved. 
 DSF is based on Dempster-Shafer Theory which is a theory of evidence combining different sources of evidence \citep{DSF2}. Let us give a brief mathematical description of DSF. 
 
DSF gets the decision profile matrix $w(x_q)$ as inputs and this matrix \citep{DSF1} is defined as follows:
\begin{equation}
w(x_q) =
 \begin{pmatrix}
  \omega_1(x_q)\\
  \omega_2(x_q) \\
  \vdots \\
  \omega_l(x_q)
 \end{pmatrix}=
 \begin{pmatrix}
  \omega_1^1(x_q) & \omega_1^2(x_q) & \cdots & \omega_1^k(x_q) \\
   \omega_2^1(x_q) & \omega_2^2(x_q) & \cdots & \omega_2^k(x_q) \\
  \vdots  & \vdots  & \ddots & \vdots  \\
   \omega_l^1(x_q) & \omega_l^2(x_q) & \cdots & \omega_l^k(x_q)
 \end{pmatrix}\,, \label{eq:dsf1}
\end{equation}
 where $\omega_i \in [0,1]^k$ and $\omega_i^j(x_q)\in [0,1]$ is a profile degree for the query image $x_q$ to belong to class $c_j$ using the classifier model $M_i$. Take note that $\sum_{j=1}^k \omega_i^j(x_q)=1$. The final decision \citep{DSF1} is given by computing the value of class support $\beta_j(x_q)$ for each class $c_j$. The class corresponding to the largest $\beta_j(x_q)$ will be considered as the final class to be affected to the input $x_q$. The computation is given as follows:
 \begin{equation}
 \beta_j(x_q)=A_j \prod_{i=1}^l \pi_j[\omega_i(x_q)]\,, \label{eq:dsf2}
 \end{equation}
 where the coefficient $A_j$ is chosen in order to ensure the expression $\sum_{i=1}^l \beta^i_j(x_q)=1$. The belief degree $\pi_j[\omega_i(x_q)]$ of the class $c_j$ obtained from $M_i$ is needed in Equation (\ref{eq:dsf2}). Its expression is given by:
\begin{equation}
\pi_j[\omega_i(x_q)]=\dfrac{\lambda^i_j(x_q) \prod_{r\neq j}\bigg (1-\lambda^i_r(x_q)\bigg)}{1-\lambda^i_j(x_q) \bigg[ 1-\prod_{r\neq j}\bigg (1-\lambda^i_r(x_q)\bigg)\bigg]}\,.
\end{equation}
The proximity $\lambda^i_j(x_q)$ between the decision template $\Lambda_j$ of class $c_j$ and the output $\omega_i(x_q)$ of query image $x_q$ is given by:
\begin{equation}
\lambda^i_j(x_q)=\dfrac{\bigg ( 1+ \| \Lambda_j^i-\omega_i(x_q)\|\bigg)^{-1}}{\sum_{r=1}^k \bigg ( 1+ \| \Lambda_r^i-\omega_i(x_q)\|\bigg)^{-1}}\;\;\text{with } \Lambda_j=\frac{1}{n_j} \sum_{z \in c_j} w(z)\,,
\end{equation}
where $\Lambda_j^i$ represents the $i^{\text{th}}$ row of $\Lambda_j$ and $n_j$ the total input data belonging to class $c_j$. As shown in Figure \ref{fig:intro}, DSF is applied twice in order to get an accurate output. Let us give an example to see how a decision can be taken through DSF.
\begin{exa}
We are given respectively the decision profile matrix and the decision templates $\Lambda_j$ in which each row $i$ represents outputs of the $i^\text{th}$ classifier and each column $j$ represents $j^\text{th}$ class, for $j=1, 2$ and $i=1,\cdots, 4$:
\begin{equation}
w(x_q)=
 \begin{pmatrix}
  0.7&0.3 \\
   0.75&0.25 \\
  0.47&0.53  \\
   0.5 & 0.5
 \end{pmatrix},\; \Lambda_1=
 \begin{pmatrix}
  0.7&0.3\\
   0.9 & 0.1 \\
  0.89 & 0.11  \\
   0.8 & 0.2
 \end{pmatrix}\;\text{and }\; \Lambda_2=
 \begin{pmatrix}
  0.3 & 0.7 \\
   0.4 & 0.6 \\
  0.3 & 0.7  \\
   0.2 & 0.8
 \end{pmatrix}
\,.
\end{equation}
From the above decision matrix $w(x_q)$, the query image belongs to class $1$ according to classifiers $1$ and $2$, it belongs to class $2$ according to the classifier 3 while we can not decide for the last classifier. This is bad news since the output of our query image depends on the classifier. Fortunately DSF solves this problem. Using Equation (\ref{eq:dsf2}), the class support of each class is given by
$\beta(x_q)=[0.011,\; 0.006]$. Since the support on the first class $0.011$ is greater than the support of the second, the query image $x_q$ belongs to the first class.
\end{exa} 

To conclude this chapter, we have seen how an image can be classified using three different classification techniques. The first technique sought to minimize the error of training; the second technique assigned a query image to a class based on similarity (closeness); the last technique transformed the problem to an optimization problem and applied KKT theorem. Since the aim of this work is to help human in decision-making as stated in our general introduction, we need to get an improved output compared to those obtained from classifiers. DSF is used for this purpose. Let us now present our results.
\chapter{Results and Discussion}
\begin{figure}[h]
        \centering
        \includegraphics[scale=0.32]{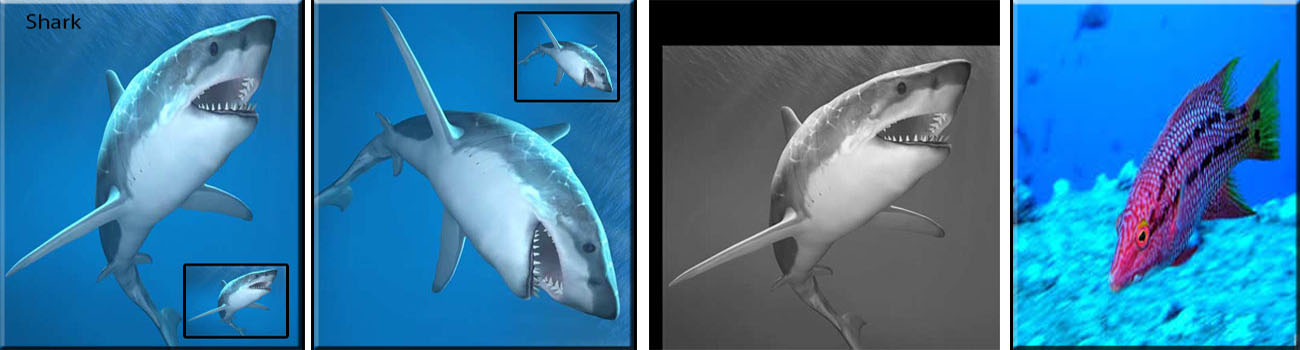}
    \caption{A shark, its geometric transformations and a fish (\href{http://www.mid-day.com/articles/us-teen-becomes-internet-sensation-by-hitching-ride-on-30ft-shark/218818}{http://www.mid-day.com/})}
    \label{fig:RES}
\end{figure}

According to the objectives assigned to this work, we should present the results on image preprocessing, feature extraction, classification and fusion.
Due to the lack of data to train our classifiers, we are able to present only the results on feature extraction. We show how the geometric transformations of an image do not influence the extraction of its features. Figure \ref{fig:RES} presents the sample of shark images used to validate our model.
\section{Results}
\begin{table}[h]
\begin{center}
\scalebox{0.7}{%

\begin{tabular}{l c*{8}{c}}
\textbf{sharks} & $\mathcal{M}_{02}\mathcal{M}_{20}$ & $\mathcal{M}_{12}^2\mathcal{M}_{20}$& $\mathcal{M}_{12}\mathcal{M}_{21}$& $\mathcal{M}_{21}^2\mathcal{M}_{02}$& $\mathcal{M}_{13}^3\mathcal{M}_{42}^3$& $\mathcal{M}_{21}^3\mathcal{M}_{02}$& $\mathcal{M}_{32}^2\mathcal{M}_{23}^2$ \\
\hline
shark & $9.0868$ & $ 4.153 $ & $1.1138$ & $4.153$ & $1.4449$ &$5.2694$ &$7.3248$ \\ 
    translated shark & $9.0868$ & $ 4.153 $ & $1.1138$ & $4.153$ & $1.4449$ &$5.2694$ &$7.3248$ \\ 
     scaled rotated shark    & $9.0868$ &$4.1528$& $1.1137$& $4.1528$ &$1.4448$ &$5.269$& $7.3241$ \\
    fish & $2.0727$ & $8.155$ &$1.1178$& $8.155$ &$1.4374$&$ 1.2551$ &$1.3135$ \\
\end{tabular}}
  \caption{CMI numerical results}
  \label{tab:CMI}
  \end{center}
  \end{table}
\begin{figure}[h]
    \centering
    \begin{subfigure}[b]{0.44\textwidth}
        \centering
        \includegraphics[width=\textwidth]{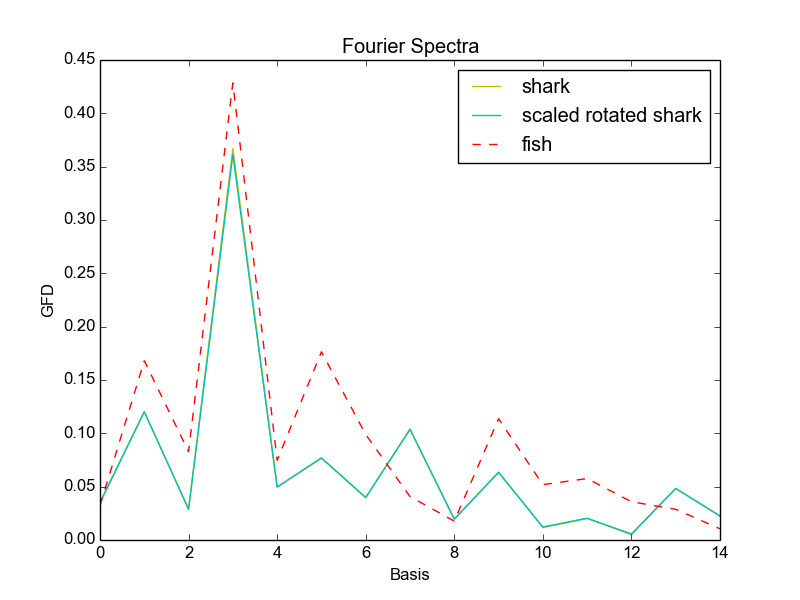}
        \caption{GFD Result}
        \label{fig:res1}
    \end{subfigure}
    \begin{subfigure}[b]{0.44\textwidth}
        \centering
        \includegraphics[width=\textwidth]{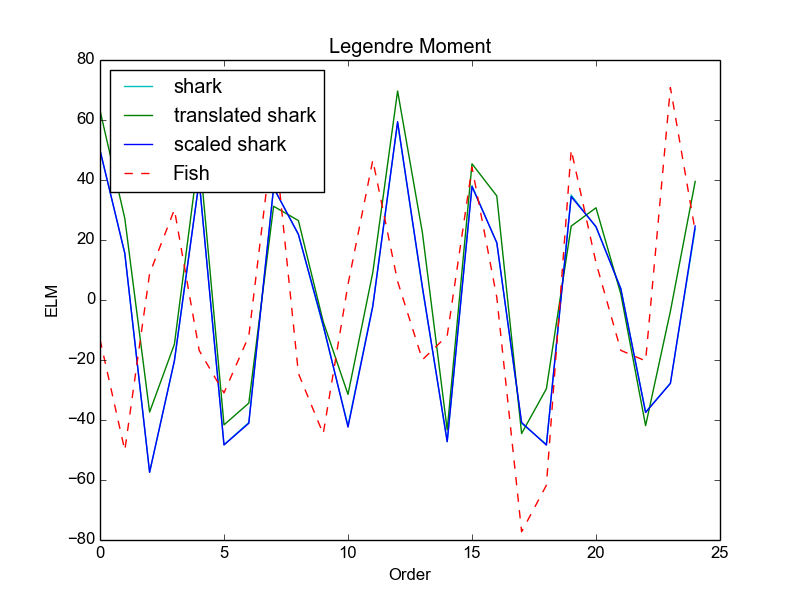}
        \caption{ELM Result}
        \label{fig:res2}
    \end{subfigure}
    \caption{Feature Extraction Results}
    \label{fig:res}
\end{figure}

Table \ref{tab:CMI} presents the results on Complex Moment. Its first row presents the different moments used to get features set; its second row presents the features obtained from the \textit{shark}; its third and fourth rows present the translated and scaled rotated shark features respectively; the last row presents the features obtained from a normal fish. Figure \ref{fig:res1} presents features obtained from the Generic Fourier Descriptor of shark and its geometric transformations as well. The same for Figure \ref{fig:res2} that presents features extracted from Exact Legendre Moment.
\section{Discussion}

We have presented in the previous section the results obtained through our programming on feature extraction. Presently, we are going to give more explanation on the results, show how they are related to the mathematical descriptions which have been developed in the previous chapters.

Concerning Feature Extraction techniques, Table \ref{tab:CMI} shows us that the shark and its different geometric transformations give approximatively the same features set in thousandth. It shows again how shark features are different compared to another object (fish). This result has confirmed the mathematical descriptions of Complex Moment made in Chapter 3. The set of features obtained from CMI are invariant under TRS and two different images do not have the same set of features.

Figure \ref{fig:res1} shows that the three curves of shark and its geometric transformations lie one another apart from the fish features presented by a dotted line. Again this result has confirmed the mathematical descriptions of Generic Fourier Transform made in Chapter 3. The set of features obtained from GFD are invariant under TRS and two different images do not have the same set of features.

As we notice from the result of Exact Legendre Moment shown in Figure \ref{fig:res2}, the translated shark gives a feature curve different to a normal shark curve while the scaled shark curve lies on the shark curve. Unfortunately, \cite{LM2} confirmed that Exact Legendre Moment is invariant under translation. This statement is contradicted by our result. The scaled variable given by $x_i=-1+i \delta x$ defined in the second section of Chapter 3 (Exact Legendre Moment) where  $\delta x =\frac{2}{n}$, requires that the two images (normal and translated) have the same size $n$, i.e.\,they must have the same variation $\delta x$ which is not the case. This is the reason of the difference of the two curves (shark and its translated form) shown in \ref{fig:res2}. Unfortunately, the same translated image worked with other feature extraction techniques such as Complex Moment as shown in Table \ref{tab:CMI}. We conclude by saying, the approximation (\ref{eq:LM5}) of the Legendre Moment is not invariant under translation, unless there is a specific image with which it works.

Concerning Image Preprocessing (IP), we did not implement its mathematical techniques. Fortunately, we have found an image package in Python called \textit{Skimage} that contains IP techniques implemented (e.g.\,Median Filter, Otsu algorithm and Random Walker algorithm), but their results are not presented in this present work due to lack of time.

After discussing on the results, we are now willing to give the overall conclusion in this coming chapter. 
\chapter{Conclusion}
We were asked throughout this work to propose an automated system supporting the actual manual shark spotter program in decision making. The motivation of this work was mainly concerned with poor weather conditions for spotters to work. We investigated that the spotter first detects a suspect object, both manual and automated systems classify whether it is a shark or not and finally a decision is taken. 

Our experiment was about achieving both theoretically and practically, the specification of the model used: to define its properties and its architecture. The architecture was built through a Bottom-Up approach, i.e.\,we have first got an idea of how the architecture had to look, motivated by previous research in Machine Learning \citep{DSF2}, and then looked for mathematical techniques that could accurately cover the model properties. Theoretically we have achieved what was expected, but an implementation lacked dataset.

Concerning the shark detection architecture that we designed (Figure \ref{fig:intro}), its first two steps are specific to the shark problem. The motivation of image preprocessing and feature extraction techniques used, are based on shark properties as described in Chapter 2. The last two steps are generic, i.e.\,any classification problem can use them. 

The interesting parts of this work were the idea of defining the relevant properties of our model and the extraction of features whose results were presented in the previous chapter. The results have confirmed the mathematical descriptions made on feature extraction techniques chiefly on Complex Moment and Generic Fourier Descriptor while the property on translation invariance was not satisfied on Exact Legendre Moment. Regarding its weakness, the automated system we proposed depends on electric power and on human attention since the images are captured by human beings (spotter).

As further work, we suggest:
\begin{enumerate}
\item Instead of using captured images to recognize sharks as stated in our general introduction, researchers can think about how to classify sharks based on tracking images.
\item It is desirable that researchers study how feature extraction functions look in terms of its properties and form. Therefore, they can focus on how to come up with their own functions based on properties they should cover depending on the problem to be treated.
\item How would hand-off be operated between beaches in the presence of shark when surfaces covered by two spotters either overlap or not: to see how the different beaches can construct a complete graph supporting each other on detection of sharks. What are the relevant spotters places in order to promote communication? In presence of a shark or school of sharks in one area, a spotter will watch their trajectory. If a shark leaves one area for another or a shark is in a surface where two beaches overlap, then the first spotter who detects a shark has to communicate with neighbouring spotters.
\item In our general introduction we saw that, with the first two flags, spotters absolve their responsibilities due to poor visibility. The present work was mainly motivated by the prevalence of the last two flags. The new idea is, in order to widen the conditions of use in the case of the first two flags, to place the cameras in different strategic positions where an object under water can easily be spotted (even attached to a hot-air balloon above the beach for example). Such positions can be chosen to decrease the frequency of either of the first two flags being deployed, and hence to improve the system as a whole.
\end{enumerate}


\chapter*{Acknowledgements}
Writing a thesis not only requires motivation, but also a lot of time. Firstly, I would like to thank my God, the Lord ALMIGHTY, the master of time and circumstances. It is through Him that I could finish this work. 

Secondly, a great thanks to my supervisors, Dr. Simukai Utete and Prof. Jeff Sanders, for giving me the means to write this work in complete serenity through their fruitful exchanges, sharp scientific opinions, always accompanied with friendly encouragement. Thanks to my tutor Martha for reading and commenting on the improvement of my work. Thanks to all AIMS staff and administration for this opportunity which has transformed my life. I would also like to thank AIMS students for their inducement. Thanks to Prof. Manya Ndjadi and Prof. Kafunda Katalay for telling me about AIMS and for recommending it to me.

I would also like to thank my parents Augustin Masakuna and Bea Indua through whom God created me. I will always be proud to have you as parents. Likewise to all my family and friends for their inducement. I would like to thank Mr. Medard Ilunga, Mr. Nico Nzau, Mr. Georges Matand, Mr. Victor Molisho, Mr. Patiano Mukishi and the rest of the ACGT for giving me this wonderful training opportunity. 

\renewcommand{\bibname}{References}
\nocite{*}
\bibliographystyle{abbrvnat}
\bibliography{references}
\addcontentsline{toc}{chapter}{References}
\end{document}